\documentclass[11pt]{article}

\usepackage[final]{acl}

\usepackage{times}
\usepackage{latexsym}

\usepackage{amsmath} 
\usepackage{amssymb}
\usepackage{wrapfig}
\usepackage{hyperref}
\usepackage{url}
\usepackage{microtype}
\usepackage{graphicx}
\usepackage{subcaption}
\usepackage{booktabs}
\usepackage{multirow}
\usepackage{xcolor}
\usepackage{colortbl}
\usepackage{listings}
\usepackage{cleveref}
\usepackage[T1]{fontenc}

\usepackage[utf8]{inputenc}

\usepackage{microtype}

\usepackage{inconsolata}

\usepackage{graphicx}

\title{Data-Centric Interpretability for LLM-based Multi-Agent Reinforcement Learning: A Case Study}
\author{%
  \textbf{John Yan}\textsuperscript{1,*}\quad
  \textbf{Michael Yu}\textsuperscript{2,*}\quad
  \textbf{Yuqi Sun}\textsuperscript{3,*}\\
  \textbf{Alexander Duffy}\textsuperscript{4}\quad
  \textbf{Tyler Marques}\textsuperscript{4}\quad
  \textbf{Matthew Lyle Olson}\textsuperscript{4}\\[0.4em]
  \textsuperscript{1}Gutenberg AI\quad
  \textsuperscript{2}MATS\quad
  \textsuperscript{3}Mindoverflow\quad
  \textsuperscript{4}Good Start Labs\\[0.2em]
  {\small \textsuperscript{*}Equal contribution.}\\
  \small{
    \textbf{Correspondence:} \href{mailto:matt@goodstartlabs.com}{matt@goodstartlabs.com}
  }
}

\newcommand{\metaautointerp}{Meta-Autointerp}

\begin{document}
\maketitle
\begin{abstract}
Large language models (LLMs) are increasingly trained via reinforcement learning in complex multi-agent environments, making it difficult to understand how behavior changes over training. Sparse Autoencoders (SAEs) have shown to be useful for data-centric interpretability. We analyze large-scale reinforcement learning runs from Full-Press Diplomacy by applying pretrained SAEs, alongside LLM-summarizer methods. We introduce Meta-Autointerp, a method for grouping SAE features into hypotheses about training dynamics. We discover fine-grained behaviors including role-playing patterns, degenerate outputs, language switching, alongside high-level strategic behaviors and environment-specific bugs. Through automated evaluation, we validate that 90\% of SAE Meta-Features are significant, and discover a reward hacking pattern where incentivized behaviors generalize into structurally similar but unrewarded actions. Through two user studies, we find subjectively interesting and seemingly helpful SAE-based features, and LLM generated hypotheses, may be harmful to human understanding. A subset of SAE-derived hypotheses are predictively useful for downstream tasks. We provide validation by augmenting an untrained agent's system prompt, improving game score by +14.2\%. SAEs and LLM-summarizer provide complementary views into agent behavior, and our framework forms a practical toolkit for monitoring and understanding LLM training. 
\end{abstract}
\begin{figure*}
    \centering
    \includegraphics[width=0.8\linewidth]{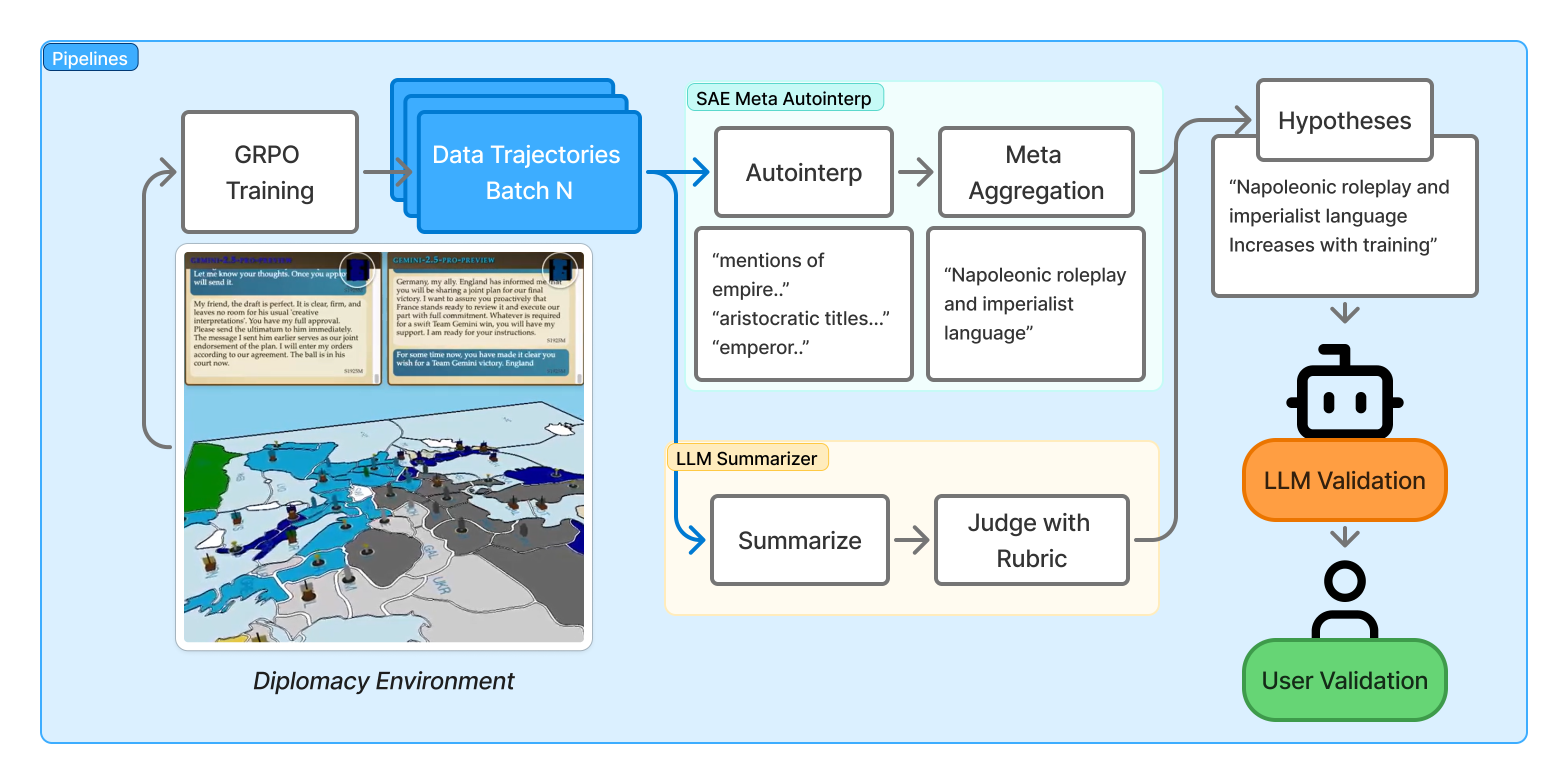}
    \caption{\small Overview of our framework. We generate hypotheses about training dynamics using LLM-based summarization and meta-autointerp over sparse autoencoder features. Hypotheses are validated using both automated LLM-based evaluations and human user studies.}
    \label{fig:overview}
\end{figure*}

\section{Introduction}
\label{sec:introduction}

Reinforcement learning (RL) is a central paradigm for training large language models (LLMs) beyond supervised learning, enabling improved reasoning and complex multi-agent coordination \citep{kaplan2020scaling, wei2022emergentabilitieslargelanguage, Shao2024DeepSeekMath, sun2024llm}. As RL environments grow more complex, understanding how and why model behavior changes during training becomes increasingly challenging. Multiple rewards and evaluation metrics often obscure qualitative differences in strategy and interaction, particularly in long-horizon or multi-agent settings \citep{duan2024gtbench}.

This limitation is especially pronounced in strategic multi-agent environments, where behaviors such as negotiation, deception, and long-term planning emerge implicitly rather than being directly supervised \citep{gandhi2023strategic, payne2025strategic2}. Prior evaluations show that agents with similar rewards can exhibit markedly different patterns of cooperation, betrayal, and strategic style \citep{duan2024gtbench, costarelli2024gamebench}, and that LLMs trained with RL may exhibit strategic misgeneralization or deceptive behavior despite strong aggregate performance \citep{greenblatt2024alignment, hagendorff2024deception, park2024ai}. 

Recent advances in mechanistic interpretability provide tools for addressing these challenges in trustworthiness \citep{elhage2021mathematical, olsson2022context}. Sparse Autoencoders (SAEs) decompose dense activations into sparse features that often correspond to human-interpretable concepts. They've even been applied in data-centric ways without requiring access to model weights \citep{movva2025sparse, jiang2025interpretableembeddingssparseautoencoders}. In parallel, LLM-summarizer methods enable scalable summarization and comparison of model behavior \citep{zheng2023judging, dubois2024llm_judge_survey}. Prior work has focused primarily on static analysis of trained models, but has rarely validated whether discovered features are accurate, reliable, or useful for downstream reasoning or intervention \citep{heap2025sparse}. 

In this work, we study training-based behavioral change in Full-Press Diplomacy, a challenging multi-agent RL environment that requires long-horizon planning and natural language negotiation \citep{duffy2025democratizing}. We analyze a large-scale RL training run comprising over 6,000 trajectories, providing a rich setting for examining how external behaviors evolve over training, without the need to analyze the trained model directly.

Figure~\ref{fig:overview} summarizes our framework. Our contributions are as follows:

\begin{enumerate}
    \item \textbf{A dual-pipeline analysis framework} that combines SAE-based feature extraction with LLM-summarization to analyze RL training runs. These approaches provide complementary perspectives, with SAEs capturing fine-grained behavioral patterns and LLM summaries highlighting higher-level strategic shifts and failure modes.
    \item \textbf{Meta-Autointerp}, a novel method for aggregating individual SAE features into coherent hypotheses about training dynamics. While many individual features are uninformative alone, aggregation yields interpretable signals tracking behavioral change over training.
    \item \textbf{Extensive validation of interpretability and helpfulness of hypotheses}. 
    Through two user studies we show that while generated hypotheses seem interesting and valid to users (in line with prior SAE evaluation work \citep{bricken2023towards}), only a subset of SAE-derived hypotheses are useful for downstream tasks and most LLM-generated hypotheses are not usable by humans. 
    To our knowledge, this is the first human study to evaluate SAE features' usefulness on a downstream task, offering a practical starting point for future work. 
\end{enumerate}
\section{Related Work}
\label{sec:related}
\paragraph{Mechanistic, Data-Centric, and Automated Interpretability}
Mechanistic interpretability aims to explain neural networks by identifying internal structures that give rise to behavior \citep{elhage2021mathematical, olsson2022context}. Sparse Autoencoders have emerged as a scalable method for decomposing dense activations into sparse, often interpretable features in language models \citep{bricken2023towards, huben2024sparse}. Prior work has shown that SAE features correspond to semantic, syntactic, and safety-relevant concepts, and large-scale efforts such as Gemma Scope provide pretrained SAEs across layers and model sizes \citep{templeton2024scaling, mcdougall2025gemmascope2tech, gao2025scaling}. More recent work has involved applying SAEs to generate hypotheses about datasets \citep{movva2025sparse, jiang2025interpretableembeddingssparseautoencoders}. Automated interpretability (autointerp) methods use language models to generate and validate natural-language explanations of SAE features \citep{bricken2023towards, paulo2025automaticallyinterpretingmillionsfeatures}. On the other hand, simple baselines can match or outperform SAEs on concept detection and probing tasks \citep{wu2025axbench,kantamneni2025probing}, and randomized baselines can obtain similar automated interpretability scores \citep{heap2025sparse,korznikov2026sanity}.

Overall, existing work for autointerp has focused only on individual features, and reviewing 10,000+ features for a typical SAE is still a manual process. And SAEs for hypothesis generation has focused on shorter contexts, limiting the applications of this technique. 

\paragraph{LLM-Based Analysis of Model Behavior}
LLM-summarizer approaches use language models to evaluate, compare, and summarize large collections of text \citep{zheng2023judging, dubois2024llm_judge_survey, sumers2025protecting}. We apply hierarchical LLM summarization to RL training trajectories, using it as a complementary analysis channel to mechanistic features: LLM summaries surface high-level strategic shifts and failure modes that are difficult to detect from activations alone \citep{xijia2025modelagnosticpolicyexplanationslarge}.

\paragraph{Diplomacy}
We develop a harness around Full-Press Diplomacy, making a challenging multi-agent benchmark requiring negotiation, long-term planning, and natural language communication \citep{duffy2025democratizing}. Prior work has focused on achieving strong performance \citep{meta2022human}, with limited analysis of how internal representations and behaviors change over training.

\section{Methods}
\label{sec:methods}

In this section, we first describe our experiment setting: the Diplomacy environment and the RL training procedure that generates our data. We then present our two complementary pipelines for extracting interpretable features from training runs. First, an SAE pipeline for fine-grained behavioral analysis, including our \textit{Meta-Autointerp} method. Second, an LLM summarization pipeline for high-level pattern discovery..

\subsection{Environment and Dataset}
\label{sec:dataset}
\paragraph{Environment}
Full-Press Diplomacy is a seven-player strategy board game, where players control major powers competing for territory. Success requires strategy, negotiation, managing relationships, and sometimes betrayal. We use a Diplomacy LLM evaluation harness \citep{duffy2025democratizing}, which provides the infrastructure for LLM agents to play complete Full-Press games.

\paragraph{LLM Diplomacy Training} 
We train LLM policies to play Diplomacy using Group Relative Policy Optimization (GRPO) \citep{Shao2024DeepSeekMath}, a policy-gradient method based on group-relative advantage estimation. Let $\pi_\theta$ denote the policy and let $\{\tau_{g,i}\}_{i=1}^K$ be a group of $K$ trajectories sampled from $\pi_{\theta_{\text{old}}}$ under identical environment conditions. Each trajectory $\tau$ is assigned a scalar return $R(\tau)$. For each group $g$, GRPO computes normalized advantages by subtracting the group mean return: $$ A(\tau_{g,i}) = R(\tau_{g,i}) - \frac{1}{K} \sum_{j=1}^K R(\tau_{g,j}) . $$ Policy updates maximize an importance-weighted policy-gradient objective: $$ \mathcal{L}(\theta) = \mathbb{E}_{\tau \sim \pi_{\theta_{\text{old}}}} \left[ \frac{\pi_\theta(\tau)}{\pi_{\theta_{\text{old}}}(\tau)} \, A(\tau) \right] . $$ The importance ratio corrects for off-policy sampling within an iteration, while group normalization bounds advantage magnitude without requiring a learned value baseline. 

We specify the Diplomacy specific actions and rewards in Section \ref{sec:implementation_details}. The overall training cost is approximately \$30,000.

\paragraph{Dataset}
We look at trajectories from two large scale training runs: one where the model successfully improved its performance, and one where it did not. Each run is 25 training steps (batches), with 8 groups per batch, and 16 trajectories per group. Each group has the same random seed for the other agents. While our sampling is constrained to only one run from each class, the massive scale of this type of training provides us with a total of 6,400 trajectories. In practical terms, we believe our data design is representative of typical training runs due to sheer cost at this scale (tens of thousands of US Dollars).

\subsection{SAE Feature Extraction Pipeline}
\label{sec:sae-pipeline}

A Sparse Autoencoder (SAE) learns to decompose a model's activation $\mathbf{x} \in \mathbb{R}^d$ into a sparse representation $\mathbf{z} = \text{ReLU}(W_{\text{enc}}\mathbf{x} + \mathbf{b}_{\text{enc}}) \in \mathbb{R}^m$ (where $m \gg d$), trained to minimize reconstruction error plus an L1 sparsity penalty. While SAEs are typically used for mechanistic interpretability of model internals, recent work \citep{movva2025sparse, jiang2025interpretableembeddingssparseautoencoders} demonstrates their utility for \emph{data-centric} interpretability: sparse features from a separate LLM serve as interpretable "tags'' that characterize text properties, identify correlations between variables, and uncover insights within datasets.

\paragraph{Activation extraction.} 
We process each trajectory's raw data (non-summarized) as follows: 
First, we tokenize the full trajectory. Next, we chunk into 1,024-token spans with a sliding window of step size 512 tokens. Third, we pass the text into Gemma 3 \cite{kamath2025gemma} to embed intermediate activations. Finally, we extract pretrained SAE activations via Gemma, taking up to 100 of the most activated features per token out of all features (262k total for the main SAE we used). This results in 6 billion activations across the corpus.

\paragraph{Feature correlation calculation.}
Our primary goal is to see how features, which represent agent behaviors, change with respect to the target variable of training step (unless specified otherwise). We calculate the correlation of each feature independently. 
We mask a given feature's activations in agent outputs ("assistant" role) only, excluding the system prompt, updates from the environment, and responses from tool calls.
Within a trajectory, we aggregate the activation values of each trajectory using one of four methods: binary (whether the feature has a non-zero activation in the sequence), max, mean, and sum.
We compute the Spearman or isotonic correlation between this aggregation (X) and our target variable (Y).
This yields 8 combinations of scoring techniques per feature (4 aggregation methods $\times$ 2 correlation types: binary/max/mean/sum aggregation with Spearman or isotonic correlation), which we find provides diverse and complementary results. We use all 8 in our analysis and run ablations over them (see Appendix~\ref{apdx:ablation-results}). Highly scoring features become candidates for validation, e.g. what agent behaviors change over training.

\subsection{\metaautointerp{}}
\label{sec:meta-autointerp}

For each candidate feature, we conduct automated interpretation (autointerp), following \citet{paulo2025automaticallyinterpretingmillionsfeatures}. We add an extra step to rate each feature 1--5 over three dimensions:
\begin{itemize}
    \item \textit{Interestingness:} How helpful is the concept for understanding agent behavior, emphasizing the multi-agent, strategic nature of the environment.
    \item \textit{Feature Coherence:} How consistent are the types of tokens a feature activates on.
    \item \textit{Context Coherence:} How consistent are the types of contexts a feature tends to activate in.
\end{itemize}

\paragraph{\metaautointerp{}} We introduce a step that groups features with similar explanations into Meta-Features.  We first filter out features that score less than a 3 on Interestingness. We then prompt an LLM to group features that activate on similar contexts and capture similar behaviors.
Our analysis finds some SAE features are not interesting or actionable in isolation, even when correctly explained by autointerp. However, these same features can reveal meaningful patterns when grouped. We generate hypotheses from these features or Meta-Features based on correlations. 

\subsection{LLM Summarization Pipeline}
\label{sec:llm-pipeline}
We compress the dataset into the final hypothesis generating LLM's context length using summarization \citep{ou2025context, chang2024booookscore}. We employ a two-stage hierarchical summarization approach inspired by \citet{sumers2025protecting}, who turn transcripts into structured interaction summaries, then summarize across summaries. To capture how training dynamics change across batches, we add an additional batch-level summarization step that preserves batch indices before final summarization. Exact prompts are shown in Appendix \ref{apdx:prompts}.

\paragraph{Stage 1: Trajectory summarization} Each trajectory ($\sim$50k tokens) is summarized to approximately 10k tokens, preserving phase structure, select tool call information, and strategic events. We use Gemini 2.5 Flash \citep{comanici2025gemini} with a prompt  emphasizing diplomatic exchanges, strategic decisions, and anomalous behaviors. 

\paragraph{Stage 2: Batch summarization} Every trajectory summary in each batch (36 summaries) is further summarized to $\sim$10k tokens, for a total of 50 batch summaries. We summarize at the batch level to ask for hypotheses for features that change over training batch.

\paragraph{Hypothesis extraction} Finally, we present the 50 batch-level summaries to Claude Opus 4.5 \cite{claude45release} to surface hypotheses about how agent behavior changes over the course of training.

\subsection{Implementation Details}
\label{sec:implementation_details}
\paragraph{Canonical Dataset}
Unless otherwise specified, we create and use a canonical set of 900 trajectories comprised of 6 randomly sampled trajectories from each group, and 6 randomly sampled groups from each batch, across the first 25 batches from one successful training run.
Each trajectory includes the full game record: all tool calls and internal thinking by the agent, responses from tool calls, and updates to the game state.

\paragraph{Trained Model} The LLM trained in the Diplomacy environment to generate the trajectories is Qwen3-235B-A22B \citep{yang2025qwen3}. We choose this model because it is the most capable open weight model offered by our third party training API provider.

\paragraph{GRPO Training}
GRPO Training uses LoRA adapters (rank 32) with importance sampling loss. For each training iteration, we collect 128 rollouts organized into groups of  16 trajectories, with a batch size of 8. The learning rate is set to 2e-4. 
We use Adam with $\beta_1$=0.9, $\beta_2$=0.95, and $\epsilon=10^{-8}$. The reward function combines per-phase and per-step components: a center delta reward of 1.0 for each supply center gained or lost, a relative position bonus of 0.2 per center above the starting count, a  small message incentive of 0.02 per message during movement phases (with a -0.05 penalty during retreat/adjustment phases), and a malformed tool penalty of -0.1. Games run for 10 phases with a maximum of 40 turns per phase. No KL penalty against the base model is applied during training. 
Actions consist of an assistant message followed by zero or more tool calls:
\texttt{get\_game\_state} (query phase and board state),
\texttt{get\_possible\_orders} (list legal unit orders),
\texttt{list\_units} (enumerate units),
\texttt{send\_message} (negotiate),
\texttt{submit\_all\_orders} (commit orders),
\texttt{write\_diary} (store notes),
\texttt{read\_diary} (retrieve notes),
\texttt{list\_rule\_files} (list references),
\texttt{cat\_rule\_file} (read rules),
and \texttt{finish\_phase} (advance the phase).

\paragraph{LLM Summarizer Model} We use Gemini 2.5 Flash \citep{comanici2025gemini} for the trajectory and batch summaries due to its balance of cost and performance, and Claude Opus 4.5 to do the final hypothesis generation over batch summaries due to its superior reasoning capabilities \citep{claude45release}.

\paragraph{SAE} We use SAEs from Gemma Scope 2 \citep{mcdougall2025gemmascope2tech} because they use the latest SAE techniques such as Matryoshka training \citep{bussmann2025learning}, and JumpReLU \citep{rajamanoharan2025jumping}. Our primary SAE is \texttt{gemma-scope-2-27b-it-resid\_post: layer\_31\_width\_262k\_l0\_medium}. We choose it because it is a canonical Gemma Scope 2 SAE, while being the largest SAE and model. We choose the middle layer of the model, which has been shown to have the most semantically interesting features \citep{templeton2024scaling, skean2025layer}.

\begin{table*}[tb]
\centering
\small
\setlength{\tabcolsep}{4pt}
\renewcommand{\arraystretch}{1}
\begin{tabular}{@{}l c p{0.45\textwidth} r r@{}}
\toprule
\textbf{Hypothesis Source} & \textbf{\# Sig.} & \textbf{Hypothesis} & $\Delta$ \textbf{Accuracy (pp)} & \textbf{$p$} \\
\midrule

\multirow{5}{*}{\textsc{SAE Meta-Features}}
  & \multirow{5}{*}{\textbf{90\%} (26/29)}
  & \cellcolor{green!15}Self-correcting mid-thought ("Wait---'', catching reasoning errors) & \cellcolor{green!15}+53.0\%* & \cellcolor{green!15}$<$1e-4 \\
  & & \cellcolor{green!8}Adopting Napoleon/imperial persona with royal titles & \cellcolor{green!8}+46.8\%* & \cellcolor{green!8}$<$1e-4 \\
  & & \cellcolor{green!15}Asking questions to gather strategic intelligence& \cellcolor{green!15}+44.0\%*& \cellcolor{green!15}$<$1e-4\\
\midrule

\multirow{3}{*}{\textsc{SAE Features}}
  & \multirow{3}{*}{\textbf{45\%} (9/20)}
  & \cellcolor{green!15}Diplomatic framing words ("alliance'', "cooperation'', "interests'') & \cellcolor{green!15}+35.0\% & \cellcolor{green!15}0.071 \\
  & & \cellcolor{green!8}Power projection vocabulary ("dominance'', "ambitions'', "threat'') & \cellcolor{green!8}+31.7\% & \cellcolor{green!8}0.092 \\
\midrule

\multirow{6}{*}{\textsc{LLM-SUMMARIZER}}& \multirow{6}{*}{\textbf{21\%} (3/14)}
  & \cellcolor{green!15}Proposing pacts while secretly planning aggression against same powers & \cellcolor{green!15}+26.7\% & \cellcolor{green!15}0.653 \\
  & & \cellcolor{green!8}Directly challenging Germany with aggressive moves & \cellcolor{green!8}+20.0\% & \cellcolor{green!8}0.653 \\
  & & \cellcolor{red!15}Confusion from phase/game-state discrepancies & \cellcolor{red!15}+14.3\% & \cellcolor{red!15}0.749 \\
  & & \cellcolor{red!8}Submitting invalid support orders (misunderstanding game rules) & \cellcolor{red!8}+12.5\% & \cellcolor{red!8}0.699 \\

\bottomrule
\end{tabular}
\caption{Early-versus-late training-stage classification using hypotheses
from three sources: LLM summaries, individual SAE features, and SAE
Meta-Features. Evaluation uses 50 sample pairs under hypothesis-random
sampling. $\Delta$ Accuracy is computed by the hypothesis minus
baseline accuracy (in percentage points). An asterisk marks a positive
difference with $p < 0.05$ by McNemar's test. Colors indicate the hypothesized
direction of change over training (\colorbox{green!15}{green} for increases,
\colorbox{red!15}{red} for decreases). Hypotheses are abbreviated for space.}
\label{tab:hypothesis-comparison}
\end{table*}

\section{Results}
\label{sec:results}

In this section, we validate hypotheses from all three sources (LLM Summarizer, SAE features, SAE Meta-Features), and discuss their results. We first conduct a user study on experts to validate our hypotheses for interpretability and helpfulness. Then we use LLM judges to validate predictive usefulness at scale, then validate with another user study for select top features. 

\subsection{User Study 1: Expert Validation on Interpretability and Helpfulness} 
\label{sec:study1}

We validate that our pipelines produce features that are both interpretable and helpful for RL practitioners, and that our meta-autointerp agrees strongly with human preferences. We gather 54 hypotheses from above 3 methods, with all 14 features from LLM-summarizer, all 20 features from SAE Meta-Autointerp and randomly sampled 20 (out of 200) from SAE Autointerp.
\begin{table}[t]
    \centering
    \small
    \begin{tabular}{@{}lcc@{}}
    \toprule
    \textbf{Method} & Helpfulness & Interpretability \\
    \midrule
    LLM Summarizer & 0.90 & 0.86 \\
    SAE Meta-Autointerp & 0.85 & 0.83 \\
    SAE Autointerp & 0.63 & 0.72 \\
    \bottomrule
    
    \end{tabular}
    \caption{\small Expert validation results (Study 1) with mean human ratings on 0-1 scale. Experts measured Helpfulness and Interpretability: if they thought a hypothesis was worth exploring further, or was easily applicable to a new sample, respectively. Meta-autointerp hypotheses outperform single feature autointerp, and LLM hypotheses obtain the highest ratings.}
    \label{tab:study1}
    \vspace{-4ex}
\end{table}

We recruited three expert developers of the Diplomacy RL environment. Each had direct experience in building and debugging the original training pipeline, but not in interpretability work. They rated all 54 hypotheses in randomized order, providing binary judgments (0 or 1) for interpretability: "Can you easily and unambiguously apply this hypothesis to new examples?'' and helpfulness: "Is this a hypothesis you would want to and could explore further?'', inspired by \citet{movva2025sparse}, and optional notes. Participants do not see which method generated each hypothesis.

Table~\ref{tab:study1} shows that SAE Meta-Features substantially outperform SAE features on both helpfulness and interpretability.  LLM-summarizer hypotheses also achieve high human ratings, likely because they capture long-horizon strategic behaviors that experts describe as "very helpful" in their notes.

We further evaluate how well autointerp's score predicts human judgment. Meta-autointerp hypotheses attain 95\% helpfulness accuracy and 90\% interpretability accuracy, compared to 75\% and 85\% respectively for autointerp. 

\subsection{Automated Hypothesis Validation}
Explainable AI research emphasizes that explanations should not only \textit{seem} interpretable to human raters, but also improve their mental models and performance on downstream tasks \citep{hoffman2023measures}. Prior human studies of RL agents have evaluated counterfactual explanations by testing whether participants can identify a flawed agent \citep{olson2021counterfactual} or recognize its learned strategy \citep{huber2023ganterfactual}. We therefore test whether our hypotheses are \textit{predictively useful}: does providing a hypothesis help an observer better distinguish between training stages? We run an automated pipeline, followed by another user study.

We collect hypotheses from the above three sources: LLM summarizer, SAE features, and SAE Meta-Features. We take one 250-token sample from the first 5 training GRPO batches and one sample from the last 5 .
Then, we ask an LLM judge about the samples twice, independently. First, by simply asking the question "Which of the two samples came earlier/later in training?''. Second, by prompting the LLM judge with the hypothesis before asking the question.

In order to source the two samples, we use two different methods.
The first is Hypothesis-random, where the positive-class sample (matching the hypothesis direction) is drawn from examples where the feature activated; the negative-class sample is drawn randomly from the corpus.
The second is Random-both, where the two samples are drawn randomly from their respective classes.

While there is a risk of evaluation circularity when LLMs both generate and judge hypotheses, we mitigate this by using three different model families as judges and validate key findings through human User Study 2 (Section~\ref{sec:study2}), which separately evaluates to what extent a subset of hypotheses can genuinely aid human classification.

Table~\ref{tab:hypothesis-comparison} provides the fraction of hypotheses that are significant (interpretable and predictively useful) from each source. We find LLM-summarizer generated hypotheses perform worst (21\% significant). Individual SAE features achieve moderate success (45\%), but grouping them into SAE Meta-Features into semantic groups via meta-autointerp substantially improves performance (90\%).

\paragraph{Discovered Features Overview}
\label{sec:features-overview}
The three hypothesis sources capture qualitatively different aspects of agent behavior. LLM summarization surfaces more global features: coordinated anti-Germany alliances, diplomatic aggression, and decreased tool-use errors. SAE features and Meta-Features capture more local features largely invisible to the summarizer: imperial roleplay, formal proposals of alliance, use of ultimatums, and self-correction in reasoning.

This difference likely reflects methodological choices: LLM summaries aggregate across full trajectories, while SAE features are extracted from only assistant turns. The complementarity suggests that combining both pipelines yields broader coverage than either alone.

\subsection{Analyzing Training Progression}
\label{sec:dynamics}
\begin{figure}[t]
    \centering
    \includegraphics[width=\columnwidth]{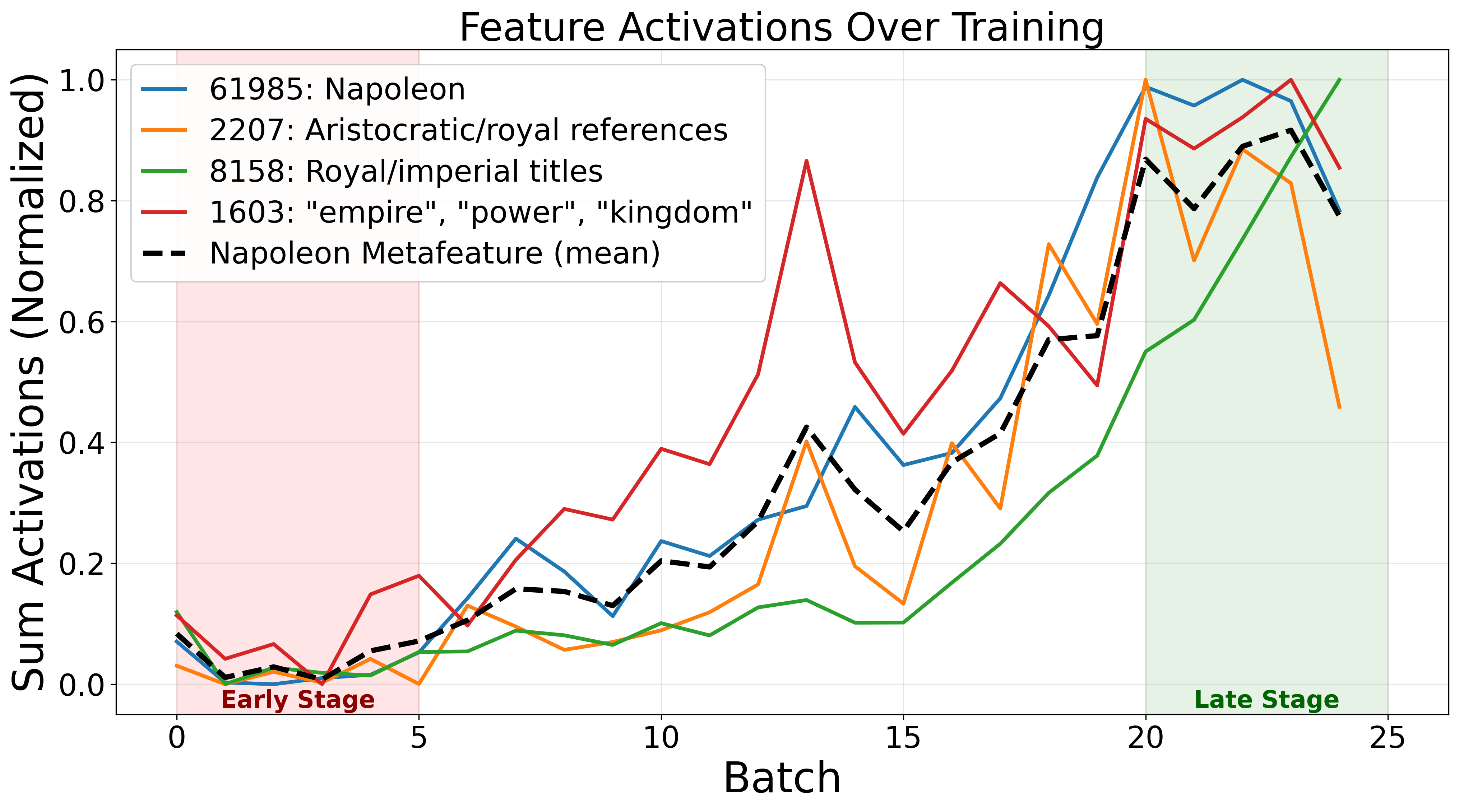}
    \caption{\small Napoleon meta-feature displayed as the mean of its component features' activations summed per trajectory, plotted over batch. Early stage and late stage classes used for predictive usefulness validation studies shown in red and green respectively.}
    \label{fig:training-dynamics}
\end{figure}
A key advantage of our SAE pipeline over LLM summarization is the quantifiable nature of activations. As shown in Figure~\ref{fig:training-dynamics}, the Napoleon meta-feature is represented by four individual features which are positively correlated with training step, showing what characteristics the agent learned over training and when. We can also examine how these features co-correlate and detect anomalies. For instance, "empire'' and "royal titles'' are positively correlated and sharply increase in step 13. We further validate the existence and coherence of these features using dense embeddings and keyword count, observing a strong correlation with activation values over training. We decompose trajectories into assistant messages and actions, generate feature-specific keywords and embeddings, then measure keyword frequencies and cosine similarity between feature labels and trajectories.

\begin{figure}[t]
    \centering
    \includegraphics[width=\columnwidth]{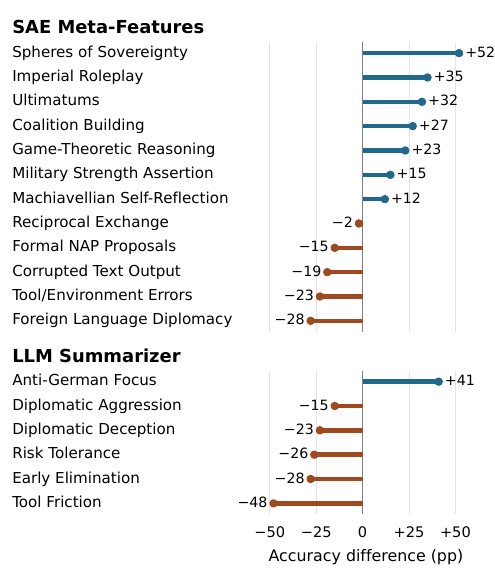}
    \caption{Human accuracy in identifying the later of two training excerpts. Each point shows the difference in accuracy, in percentage points, between the hypothesis-assisted condition and the baseline. Accuracy was higher than baseline for 7 of 12 SAE Meta-Features and 1 of 6 LLM-summarizer hypotheses.}
    \label{fig:study2-human-results}
\end{figure}


\subsection{User Study 2: Predictive Usefulness}
\label{sec:study2}

We aim to demonstrate that our features are not just \emph{predictively useful} to LLMs, but for human users. Our goal is to see if our framework can help practitioners distinguish between samples from different conditions, for example early versus late in training. Unlike previous work on SAE feature interpretability \citep{jiang2025interpretableembeddingssparseautoencoders, movva2025sparse}, we show features are useful for downstream tasks instead of relying on subjective qualitative ratings. We measure whether providing a hypothesis improves accuracy on a paired early-vs-late training classification task using identical text span pairs with and without hypotheses. Performance is compared using a paired McNemar test over per-pair correctness, using accuracy difference in mean accuracy between conditions. We conduct a study on compensated volunteers by collecting 277 responses across 18 hypotheses from 32 participants. Participants spend on average 10 minutes responding, 3 of which randomly receive a payout compensation of 25 dollars. 

\paragraph{Human Results}
\label{sec:study2-human}
Results in Figure~\ref{fig:study2-human-results} show high variance in hypothesis predictive usefulness. Several SAE generated hypotheses substantially improve early vs late classification accuracy, including Spheres of Sovereignty (+52\%) and Imperial Roleplay (+35\%). Sparsity of feature activations is a major limitation. In Foreign Language Diplomacy (-28\%), several sampled span pairs have no instances of the behavior. Several SAE features are quite subtle and difficult to distinguish by humans, especially within the short survey time and short text spans. While SAE features are subjectively interesting, our empirical results suggest that they are much harder for users to use on a downstream task. Among LLM generated hypotheses, only Increasing Anti-German Focus (+41\%) increase accuracy. We hypothesize both LLM and SAE generated hypotheses are more useful with larger sample sizes and context, however LLM features are disproportionately impacted as they capture higher level features depending on longer context.

\section{Analysis}
\label{sec:analysis}

In this section, we explore to what extent the generated hypotheses can be useful outside of subjective and predictive human interpretability. We find a few diverse applications for how our framework can be leveraged as part of a toolbox. We leave comparisons to other tools (perplexity measures, KL-divergence tracking, linear probe fitting, etc.) for future work. 

We first design an experiment to show how the discovered hypotheses can be actionable: improving the score in diplomacy in an off-the-shelf model. We then run two case studies into large scale training runs, showing that if SAE-based features had been available, unintended behavior (poor performance and reward hacking) could have been identified early in runs. 

\subsection{Hypothesis-Guided Prompt Optimization}
\label{sec:prompt-interfere}
We want to test whether our collected hypotheses were \textit{actionable} and \textit{effective}, not just interpretable. If the hypotheses capture behaviors that are actually effective in the Diplomacy environment, we should be able to prompt an agent to do those behaviors to see if performance improves.

We run 20 games of non-training diplomacy, ten under control condition and ten for intervention condition. We use the base model from our early experiments, the Qwen3 \citep{yang2025qwen3} model, which has not been trained in the environment. In the control condition, the France agent is given only the standard system prompt. In the intervention condition, the France agent receives the same system prompt, augmented with 10 selected behavior hypotheses from our SAE Meta-Features pipeline, such as Coalition Building and Reciprocal Exchange.  The full intervention prompt and categories with examples is provided in Appendix~\ref{apdx:intervention_prompt}.

While the experiment design (20 games, bundling 10 hypotheses), we find that under the intervention condition, the agent achieves a significantly higher average game score. The agent with hypotheses scores games with a mean of $43.65 \pm 8.06$ compared to $38.20 \pm 2.24$ for the baseline, which corresponds to an average improvement of $+5.45$ points, or $+14.2\%$. Despite a low sample count, a two-sided $t$-test indicates that this difference is statistically significant ($t=2.91$, $p=0.006$). 
Additionally, even though variance increases in the intervention condition, the higher mean suggests that hypothesis optimized prompts increase performance overall. These results demonstrate that the outputs of our framework are useful for downstream tasks

\begin{figure}[tp]
    \centering
    \includegraphics[width=\linewidth]{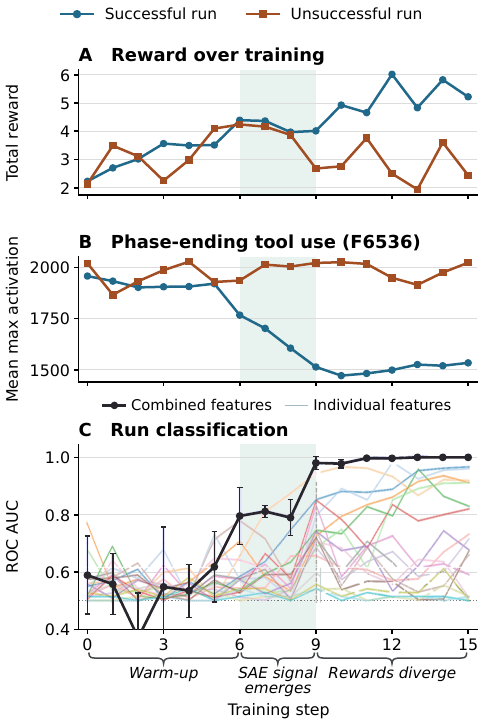}
    \caption{Behavioral divergence between two Diplomacy training runs.
    \textbf{(A)} Total reward in the successful and unsuccessful runs.
    \textbf{(B)} Mean maximum activation of SAE feature F6536, associated
    with the \texttt{finish\_phase} tool call.
    \textbf{(C)} ROC AUC for classifying which run produced a trajectory,
    using each SAE feature individually (faint lines) or all 20 features
    together (dark line with error bars). After warm-up, feature activation
    begins to separate the runs at step~6, while their rewards remain
    similar. The reward curves diverge at step~9.}
    \label{fig:feature_divergence}
\end{figure}

\subsection{Case Study 1: Early Identification of Bad Training Runs}
One of our GRPO training runs in the Full Press Diplomacy environment does not improve, which we investigate in comparison with the successful run. As shown in Figure~\ref{fig:feature_divergence}A, we analyze these two runs: a \emph{good} run that continues to improve and a \emph{bad} run whose performance plateaus. Before training batch 9, both runs achieve comparable reward, making them indistinguishable under standard learning curves. However, SAE analysis uncovers a sharp divergence at this stage. Using the top 20 automatically interpreted SAE features from each checkpoint, we fit a linear probe to classify whether a trajectory comes from either run, calculating the false positive rate vs true positive rate robustly regardless of class imbalance. As shown in Figure~\ref{fig:feature_divergence}C, probe performance rapidly increases starting batch 6 with a combined  AUC $\geq$ 0.8, and reaches near-perfect AUC after step 9. The combined AUC is calculated from logistic regression trained on standardized activations of 20 selected SAE features, evaluated via 5-fold CV at each training step. 

Further inspection shows that this signal is driven by a single feature. In the bad run, as shown in Figure~\ref{fig:feature_divergence}B, this feature remains approximately constant throughout training, while in the good run it diverges sharply starting batch 6. Manual inspection reveals that this feature corresponds to correctly using the tool to end a phase of the game. The good run learns to use this, whereas the bad run fails to do so. The divergence in feature activation occurs at batch 6, whereas the divergence in reward occurs at batch 9, meaning our method provides an earlier, more interpretable signal for distinguishing good and bad runs.

\subsection{Case Study 2: Reward Hacking}
\label{sec:coherence}

When analyzing good training runs with our framework, we find a variety of SAE features, one of which captures the agent behavior of sending duplicate messages: not surprising as our reward function (Section~\ref{sec:implementation_details}) gives +0.02 per message during movement phases. While manually investigating these behaviors, we are surprised to find the agent also starts writing duplicate diary entries, a behavior similar to duplicate message sending but does not result in any reward.

\begin{figure}[t]
    \centering
    \includegraphics[width=\columnwidth]{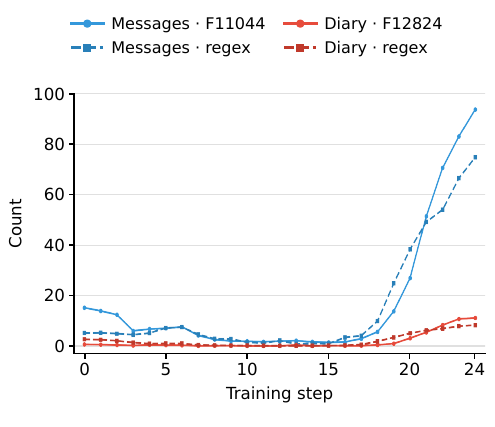}
    \caption{Duplicate messages and diary entries during the successful training run. Solid lines show counts from SAE features F11044 and F12824. Dashed lines show counts from regular-expression matching. Both behaviors became more frequent late in training, although only messages earned a reward.}
    \label{fig:rh_fig}
\end{figure}

In Figure~\ref{fig:rh_fig}, we show that two SAE features present high correlation, and their accuracy is validated by Regex keyword matching. This highlights an unexpected RL training pattern: reward hacking behaviors can overflow into structurally similar but unrewarded actions. Our framework provides a practical example of detecting previously hidden patterns in RL training dynamics.

\section{Conclusion}
\label{sec:conclusion}
We present a framework for interpreting LLM training in complex multi-agent RL environments. Our SAE Meta-Autointerp and LLM-summarizer pipelines reveal complementary insights. Through user studies and downstream validation, we find that not all interpretable features are useful. Certain features that appear helpful are counterproductive when used by humans, but the right features can predict training dynamics and guide interventions.

\section*{Limitations}
Our analysis is constrained to a single environment with limited training runs. While we observe strong correlations between features and outcomes, causal relationships and model internals are largely untested, and our intervention experiment combined multiple features rather than ablating each individually. It is also clear that SAE and AutoInterp robustness also depends on choice of LLM, and newer versions of LLM architectures may not lend themselves to SAEs or AutoInterp. Our intervention experiment validates that discovered features are actionable at inference time, while we leave out the stronger test of using these features to monitor and intervene on live training runs for future work. Future work includes intervention experiments at scale, SAEs with longer context windows, and validation across different environments.

\newpage
\bibliography{02biblio}
\appendix

\section{Experiment Setup: Details}
\label{apdx:experiment_setup_details}
\paragraph{Windowing Samples}
The samples we pass to the judge are 250 tokens. We get samples from 4 sources:
\begin{itemize}
    \item \textbf{LLM-Summarizer hypotheses}: In our summarization pipeline, we have a judge get citations from the raw transcripts that illustrate the hypotheses. These can be of arbitrary length. The 250-token window is then centered around this citation.
    \item \textbf{SAE feature hypotheses}: We randomly select top activating examples from a given feature, and center a 250-token window around the top activation.
    \item \textbf{SAE meta-feature hypotheses}: We randomly select top activations from any features in a given meta-feature, and center a 250-token window around the top activation.
    \item \textbf{Random}: We randomly center a 250-token window around any token in an assistant's message.
\end{itemize}

\paragraph{Multiple Judges}
To assess whether pooling across LLM judges is appropriate, we computed Fleiss' $\kappa$ for inter-rater agreement on each sample pair. At baseline (no hypothesis), agreement was poor ($\kappa = 0.04$), indicating models relied on idiosyncratic heuristics. With the hypothesis provided, agreement rose to substantial ($\kappa = 0.61$), with 71\% unanimous agreement and 97\% accuracy when all three models concurred. This convergence suggests the hypothesis provides a genuine shared signal rather than amplifying model-specific biases, justifying pooled analysis across judges.

 \begin{figure}
    \centering
    \includegraphics[width=\linewidth]{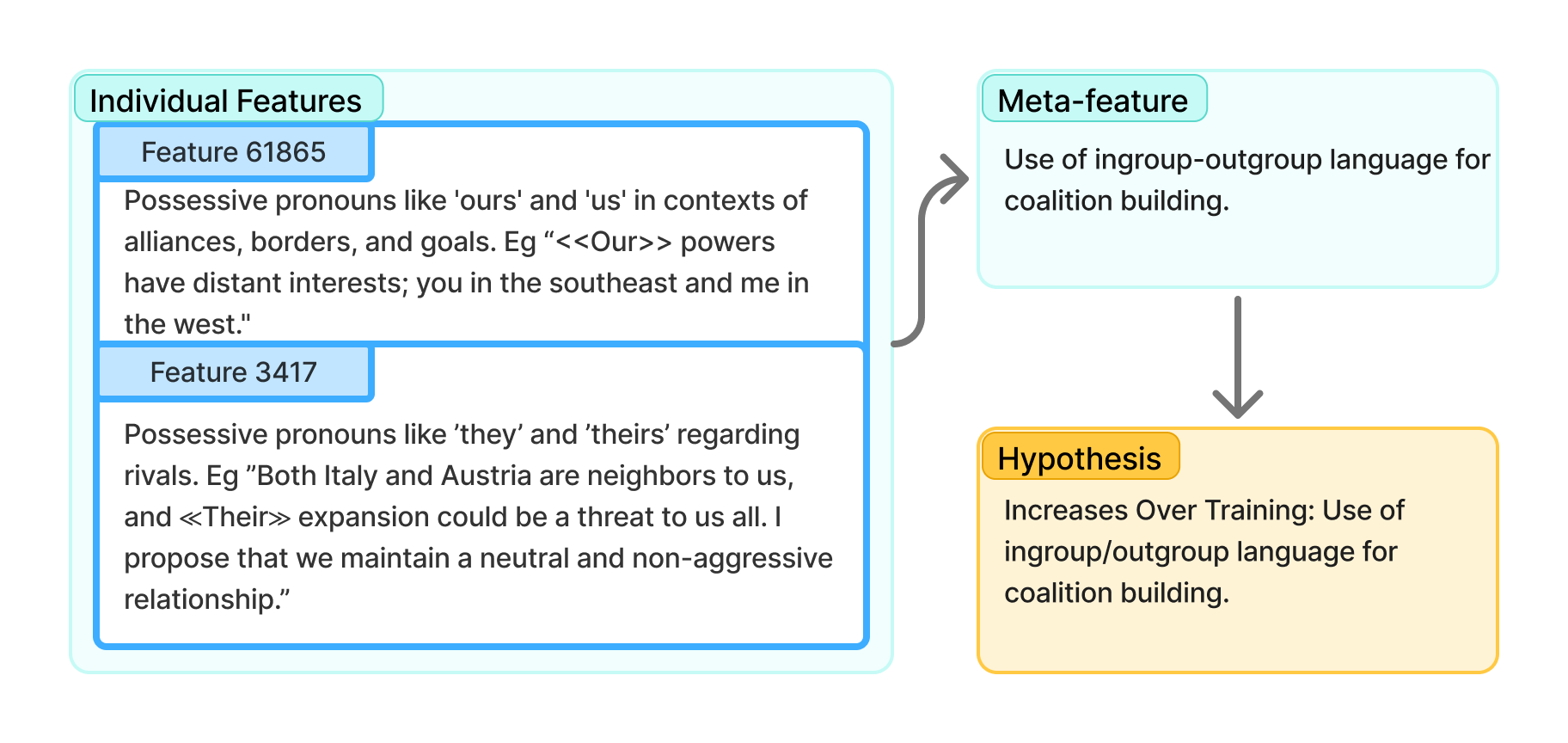}
    \caption{The hypothesis generation workflow. Meta-Features provide insight that single features don't. Note: hypotheses can also be generated from individual features. Maximally activating tokens in a sequence are delimited like $\ll$so$\gg$, following convention \cite{paulo2025automaticallyinterpretingmillionsfeatures}}
    \label{fig:hypothesis_example}
\end{figure}

\paragraph{Random Sampling}
We tested two sampling strategies: \textbf{hypothesis-random} (target sample drawn from high-activation examples, non-target drawn randomly) and \textbf{random-both} (both samples drawn randomly from their respective classes). Table~\ref{tab:sampling-modes} compares performance.
\begin{table}[h]
\centering
\caption{Accuracy difference (\%) by hypothesis source and sampling mode}
\label{tab:sampling-modes}
\begin{tabular}{lcc}
\toprule
& \textbf{hypothesis-random} & \textbf{random-both} \\
\midrule
\textbf{LLM} & +8.3 & +8.3 \\
\textbf{Features} & +8.0 & +7.0 \\
\bottomrule
\end{tabular}
\end{table}

Absolute accuracy difference is consistent across modes ($\sim$8\%). We choose to report hypothesis-random throughout the paper for two reasons: (1) consistency with our human user study, which necessarily uses activation-guided sampling to ensure participants see relevant examples, and (2) random-both requires higher sample volumes to achieve reliable signal, increasing computational cost.

\paragraph{Judge Prompt}
Baseline prompt
\begin{lstlisting}
You are examining outputs from an LLM playing the strategy game Diplomacy. The agent interacts with the environment through tool calls, in order to move troops, communicate with players, or poll the game state.

{type_specific_question}

Return ONLY a JSON object in this exact format:
{{
    "answer": "A or B",
    "explanation": "brief 1-2 sentence explanation"
}}

### Sample A:
{sample_a}

### Sample B:
{sample_b}
\end{lstlisting}

Hypothesis prompt
\begin{lstlisting}
You are examining outputs from an LLM playing the strategy game Diplomacy. The agent interacts with the environment through tool calls, in order to move troops, communicate with players, or poll the game state.

{type_specific_question}

Hypothesis to consider: {hypothesis}

Return ONLY a JSON object in this exact format:
{{
    "answer": "A or B",
    "explanation": "brief 1-2 sentence explanation"
}}

### Sample A:
{sample_a}

### Sample B:
{sample_b}
\end{lstlisting}
\section{Ablation Studies}
\label{apdx:ablation-results}

\begin{figure}
    \centering
    \includegraphics[width=0.75\linewidth]{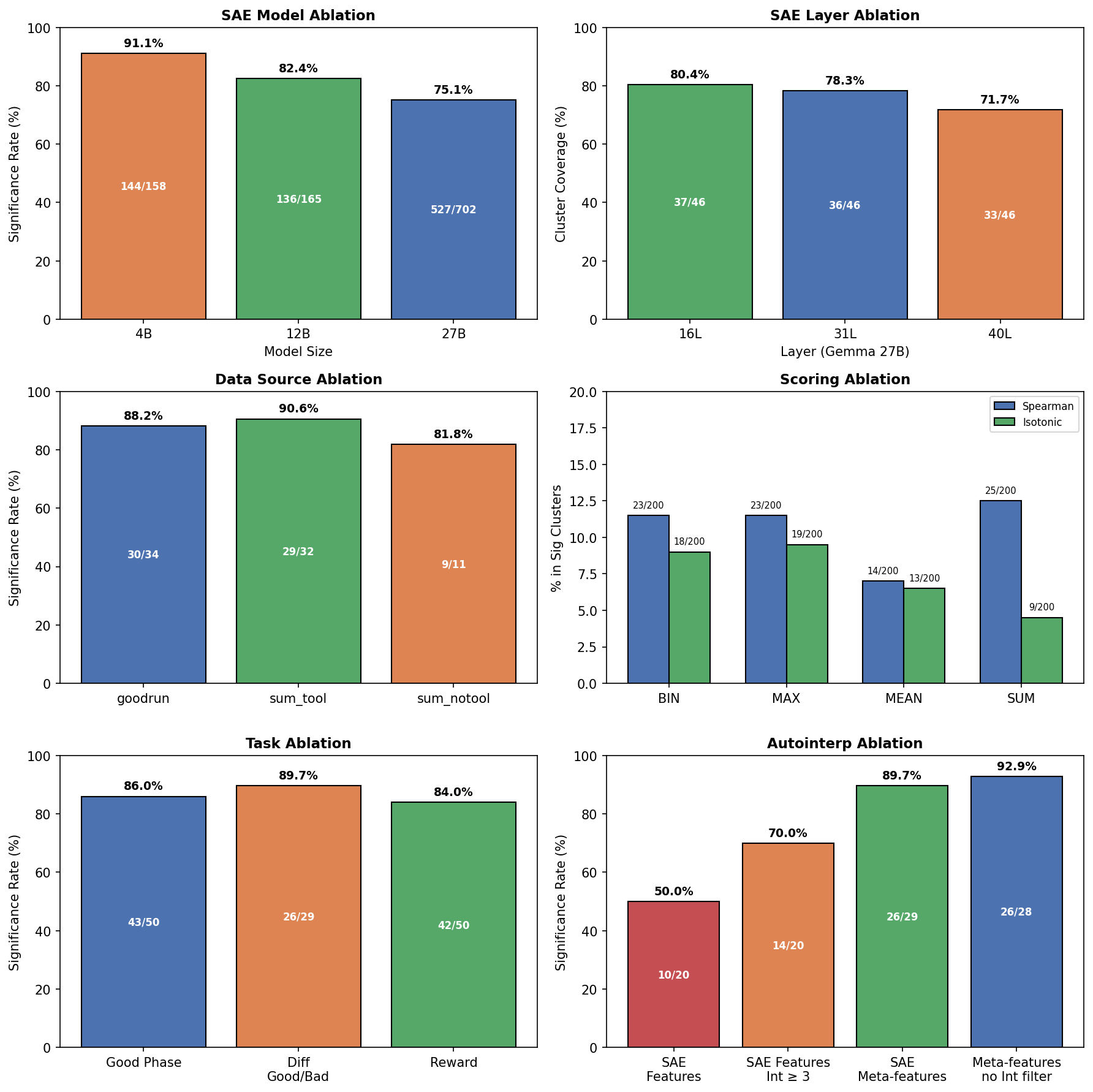}
    \caption{Top left: ablations across model size. Top right: ablations across model layer. Middle left: ablations across data source. Middle right: ablations across aggregation and correlation method. Bottom left: comparisons across correlation target. Bottom right: ablations for Interestingness filter.}
    \label{fig:ablation-bars}
\end{figure}

We conducted multiple ablation experiments to validate design choices in our hypothesis generation pipeline. Figure~\ref{fig:ablation-bars} summarizes results.

\subsubsection{Multi-SAE Grouping}
We can apply Meta-Autointerp to more than just a single SAEs features. We use the same primary model, Gemma 3 27B-IT 31L 262K canonical, as featured in ~\ref{tab:hypothesis-comparison}.

On the same data and correlation target, we also extract the features from these SAEs:
\begin{itemize}
    \item Gemma 3 27B-IT 16L 262K canonical
    \item Gemma 3 27B-IT 40L 262K canonical
    \item Gemma 3 12-IT 24L 262K canonical
    \item Gemma 3 4B-IT 17L 262K canonical
\end{itemize}

We also apply the SAE for Gemma 3 27B-IT 31L 262K canonical on 2 additional datasets:
\begin{itemize}
    \item sum\_notool: the summarized trajectories from ~\ref{sec:llm-pipeline}
    \item sum\_tool: the same summarized trajectories from sum\_notool, but with all agent tool calls preserved verbatim
\end{itemize}

 We can sweep across the middle SAE layer across model sizes, SAE layers on the 27B model, and also the Gemma 3 27B-IT 31L 262K SAE across the raw data and 2 different kinds of summarization. We feed all of these features into a single Meta-Autointerp run, and validate. We find that the significance rate is 94\%, only marginally higher than the run using a single SAE and data source.

 We can attribute which models, layers, and data sources uniquely contributed to significant Meta-Features. We find that across all of them, the difference in significance rate is not high.

\subsubsection{Correlation Scoring Ablation}
Similarly take the Meta-Features from our single canonical SAE and data source, to find what proportion of features provided by each scoring methodology ended up being significant. We find that sum + spearman correlation gives the highest rate, but the methods tend to provide qualitatively different features, so continue to pool them.

\subsubsection{Task Target: Phase vs Reward vs Diff}
In addition to our primary target: batches over a successful run, we also perform 2 more Meta-Autointerp runs:
\begin{itemize}
    \item \textbf{Diff Good/Bad:} After getting the aggregated value per feature per trajectory in the successful and unsuccessful runs, we subtract the means of (goodrun-badrun). We then score this with regards to batch step to measure the divergence between a successful and unsuccessful run.
    \item \textbf{Reward:} We have a reward with each trajectory, provided by the environment. We use the same aggregation and correlation scoring methods, but with regards to the reward value. This is to find behaviors that are correlated with highly/lowly rewarded trajectories.
\end{itemize}

We find the significance rate across all 3 tasks is about the same.

\subsubsection{Autointerp Interestingness Filtering}
Our autointerp pipeline assigns each feature an Interestingness score (1--5). Typically, we filter out features that score less than 3 on Interestingness before grouping through Meta-Autointerp. We wanted to see:
\begin{itemize}
    \item How significant were single features after being filtered for Interestingness $\geq$ 3?
    \item How significant were Meta-Features without being filtered for Interestingness $\geq$ 3?
\end{itemize}
We found that for the former, significance rate improved, but was still not as significant as Meta-Features. And for the latter, significance rate was about even.

\section{User Study 2: Details}
\begin{table*}[t]
  \caption{Hypothesis validation results from human evaluation.}
  \label{apdx:user_study_2_table}
  \begin{center}
      \begin{tabular}{lccc}
        \toprule
        Hypothesis & Accuracy & $\Delta$ Accuracy (pp) & Source \\
        \midrule
        Spheres of Sovereignty & 1.00 & +0.52 & SAE \\
        Increasing Anti-German Focus & 0.89 & +0.41 & LLM \\
        Imperial Roleplay & 0.83 & +0.35 & SAE \\
        Ultimatums & 0.80 & +0.32 & SAE \\
        Coalition Building & 0.75 & +0.27 & SAE \\
        Game-Theoretic Reasoning & 0.71 & +0.23 & SAE \\
        Military Strength Assertion & 0.62 & +0.15 & SAE \\
        Machiavellian Self-Reflection & 0.60 & +0.12 & SAE \\
        Reciprocal Exchange & 0.46 & -0.02 & SAE \\
        Formal NAP Proposals & 0.33 & -0.15 & SAE \\
        Increasing Diplomatic Aggression & 0.33 & -0.15 & LLM \\
        Corrupted Text Output & 0.29 & -0.19 & SAE \\
        Tool/Environment Errors & 0.25 & -0.23 & SAE \\
        Increasing Diplomatic Deception & 0.25 & -0.23 & LLM \\
        Increasing Risk Tolerance & 0.22 & -0.26 & LLM \\
        Decreasing Early Elimination & 0.20 & -0.28 & LLM \\
        Foreign Language Diplomacy & 0.20 & -0.28 & SAE \\
        Decreasing Tool Friction & 0.00 & -0.48 & LLM \\
        \bottomrule
      \end{tabular}
  \end{center}
\end{table*}

\label{apdx:user_study_2_details}
\paragraph{Evaluation and Statistics.}
We measure whether providing a hypothesis improves accuracy on a paired early-vs-late training classification task using identical text span pairs with and without hypotheses. Performance is compared using a paired McNemar test over per-pair correctness, with $\Delta$ Accuracy defined as the difference in mean accuracy between conditions.

\paragraph{Design.} We evaluate whether feature-derived hypotheses improve class prediction between early and late training samples. We consider hypotheses from 6 LLM features and 13 SAE Meta-Features. Each task asks: “Which sample occurred later in training?” and presents a pair of text spans, one sampled from early training and one from late training.

In the non-baseline condition, we additionally provide a hypothesis such as “increases with training: aggression.” In the baseline condition, no hypothesis is shown.

For each hypothesis, we sample 5 text pairs, yielding 90 tasks with hypotheses and 90 matched baseline tasks using identical span pairs. Each participant completes 10 questions total. We use a within subject study design, participants see 5 baseline tasks followed by the same 5 pairs with hypotheses, presented in randomized order. We cap participation at 10 questions per user to limit hypothesis memorization.

\paragraph{Metrics.}
Let $y_i^B, y_i^H \in \{0,1\}$ denote baseline and hypothesis correctness for span pair $i$.
$$
\text{Accuracy}_B = \frac{1}{N}\sum_i y_i^B
$$

$$
\text{Accuracy}_H = \frac{1}{N}\sum_i y_i^H
$$
\[
\Delta \text{Accuracy} = \text{Accuracy}_H - \text{Accuracy}_B
\]

\paragraph{McNemar Test.}
We construct a $2\times2$ contingency table over paired outcomes:
\[
\begin{array}{c|cc}
 & y^H=1 & y^H=0 \\\hline
y^B=1 & a & b \\
y^B=0 & c & d
\end{array}
\]
and apply an exact McNemar test on the discordant counts $(b, c)$ to test whether hypotheses change classification accuracy.

\section{User Study 1: Details}
\label{apdx:user_study_1_details}

\begin{figure*}
    \centering
    \includegraphics[width=\linewidth]{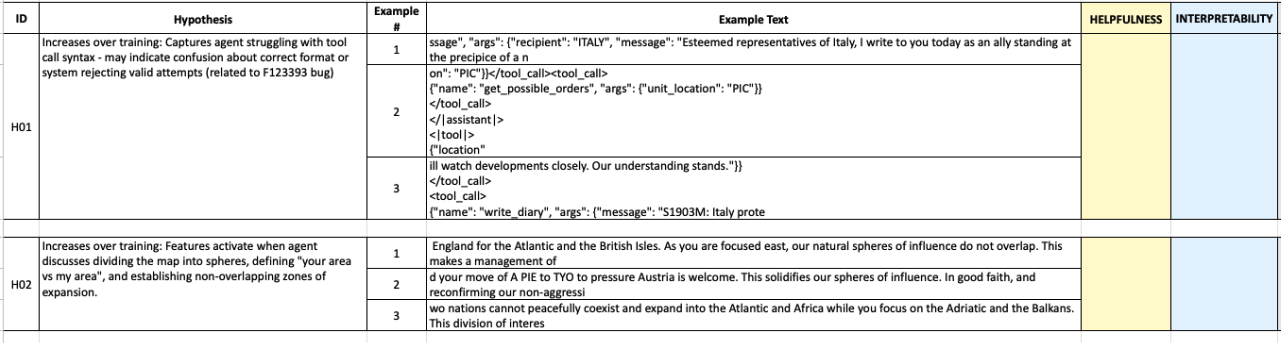}
    \caption{   Screenshot of user study 1 interface}
    \label{fig:user-study-1-interface}
\end{figure*}

\paragraph{Participants.}
We recruited three expert developers of the Diplomacy RL environment. All participants had direct experience building and debugging the training pipeline.

\paragraph{Materials.}
Each participant rated 54 hypotheses: 14 from the LLM-Summarizer pipeline, 20 individual SAE features (randomly sampled from top-200 correlated features), and 20 SAE Meta-Autointerp features (created by first filtering single feature Autointerp interpretability scores $\geq 3$ and grouped). Hypotheses were presented in randomized order.  Each hypothesis included an id, explanation and 3 examples. We did not show participants what kind of feature (LLM-summarizer feature, SAE feature or SAE meta-feature) the hypothesis is generated from.

\paragraph{Rating Rubric.}
Participants provided binary ratings (0 or 1) on two dimensions:

\textbf{Helpfulness:}
\begin{quote}
Does this hypothesis help you understand how model behavior changes over training? If you were studying this dataset, is this a hypothesis you would want to and could explore further? Rate 1 if yes, 0 if no or only a little.

\textit{Good example:} "Increases over training: Agent uses strategic deception to mislead opponents about intentions''

\textit{Bad example:} "Increases over training: Token appears at end of sentences''
\end{quote}

\textbf{Interpretability:}
\begin{quote}
When you read the hypothesis, is it clear what it means? Can you easily and unambiguously apply it to new examples? Rate 1 if yes, 0 if no or it would often be subjective.

\textit{Good example:} "Decreases over training: Agent explicitly states unit locations using coordinate format''

\textit{Bad example:} "Increases over training: Agent shows complex reasoning patterns''
\end{quote}

\paragraph{Interface.}
Ratings were collected via a Google spreadsheet interface (\Cref{fig:user-study-1-interface}).

\section{Feature Distributions}
\label{apdx:distributions}

\begin{figure}
    \centering
    \includegraphics[width=\linewidth]{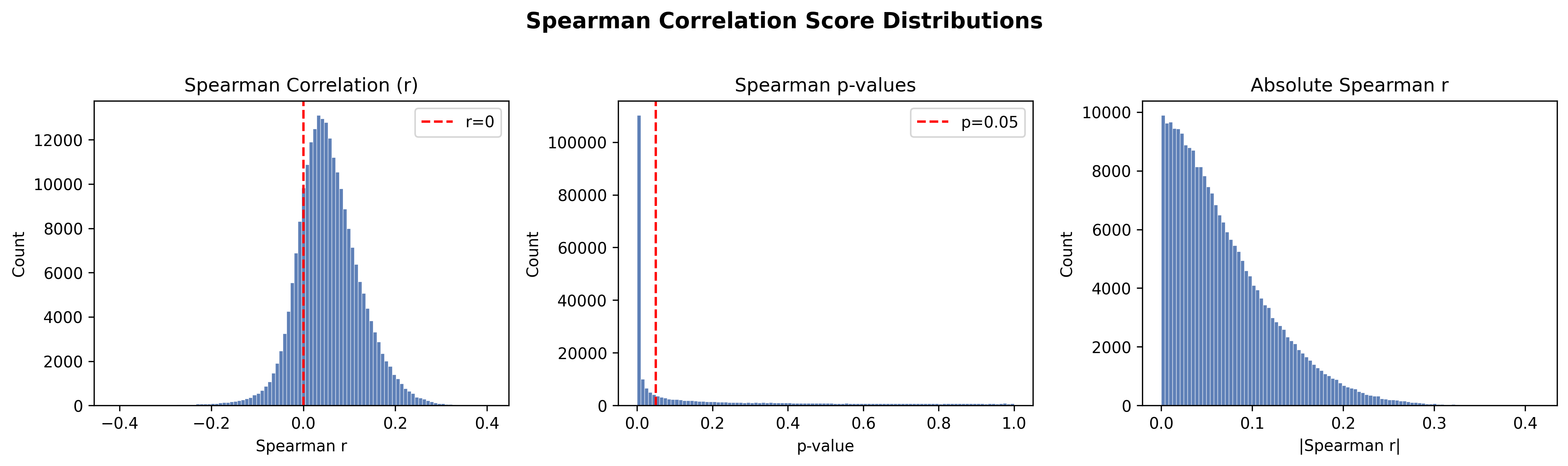}
    \caption{   Distribution of Spearman correlations between SAE features and training batch. Around 60\% of features are positively correlated with training batch}
    \label{apdx:fig:spearman_distribution}
\end{figure}

We plot the distribution of spearman correlation scores of feature activations against training step. The scores are approximately normally distributed with a mean above 0, suggesting a majority of features are positively correlated with training (See \Cref{apdx:fig:spearman_distribution})

\section{Embeddings and Keyword Validation}
\label{apdx:training-dynamics}

To further validate the coherence and consistency of features, we use keyword counting and embeddings. We parse out all assistant messages, write diary, and send message tool calls (actions) from each trajectory to create embeddable documents and have a more fine-grained unit of analysis than full trajectories. 

\subsection{Keyword Counting}
For each feature, we generate 20-50 keywords with Claude Sonnet 4.5 We then sum the total occurrences of each keyword in all actions in each trajectory, taking the mean over each batch and normalizing over actions length and count.

\subsection{Embeddings}
For embeddings, we use OpenAI text-embedding-large to generate 1536 dimensional vectors for each action in each trajectory, and each feature label. We then calculate the cosine similarity between the feature label and each document, averaging cosine similarity over each batch.

\begin{figure}
    \centering
    \includegraphics[width=\linewidth]{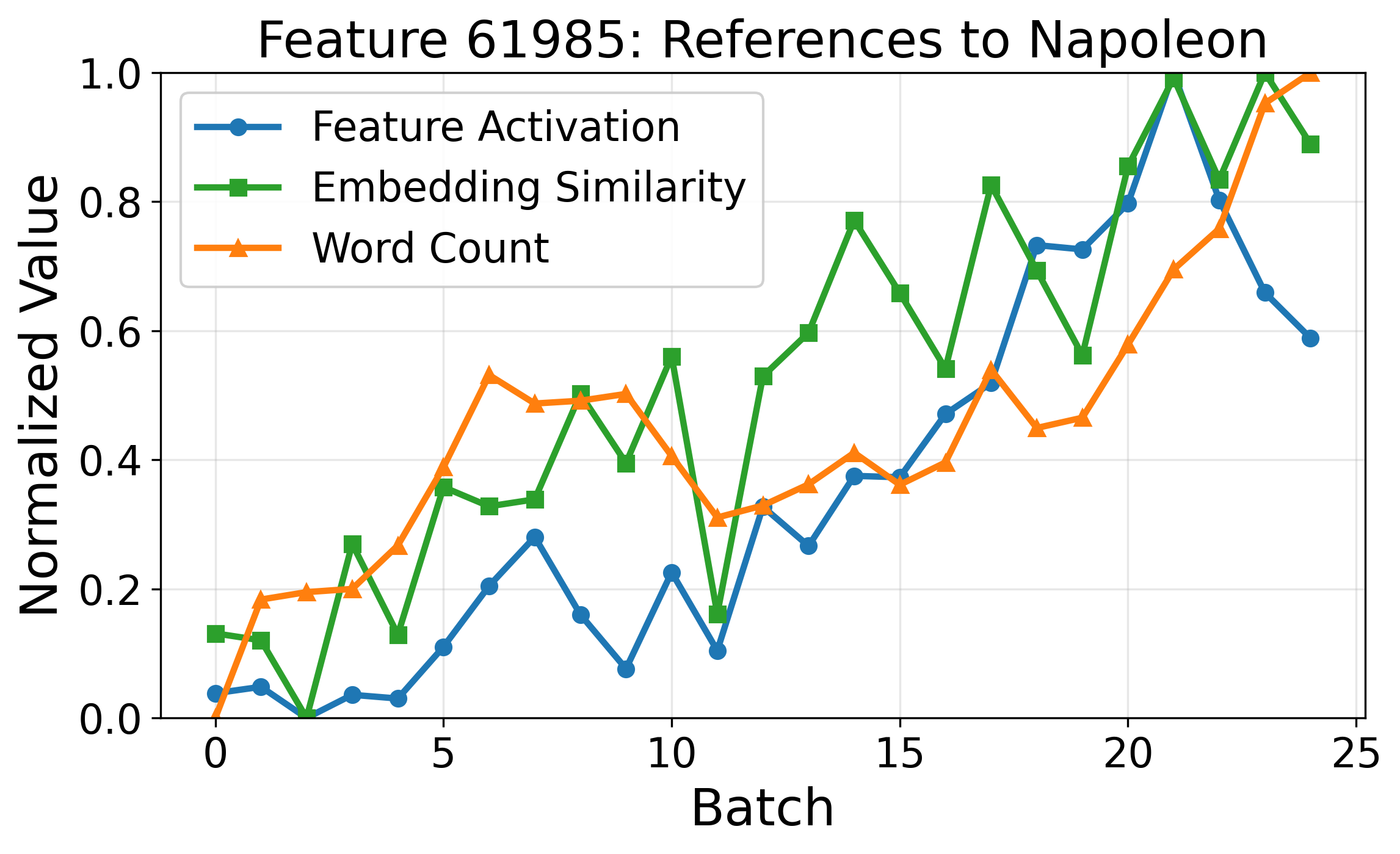}
    \caption{References to Napoleon}
    \label{apdx:keyword_embeddings_napoleon}
\end{figure}

\begin{figure}
    \centering
    \includegraphics[width=\linewidth]{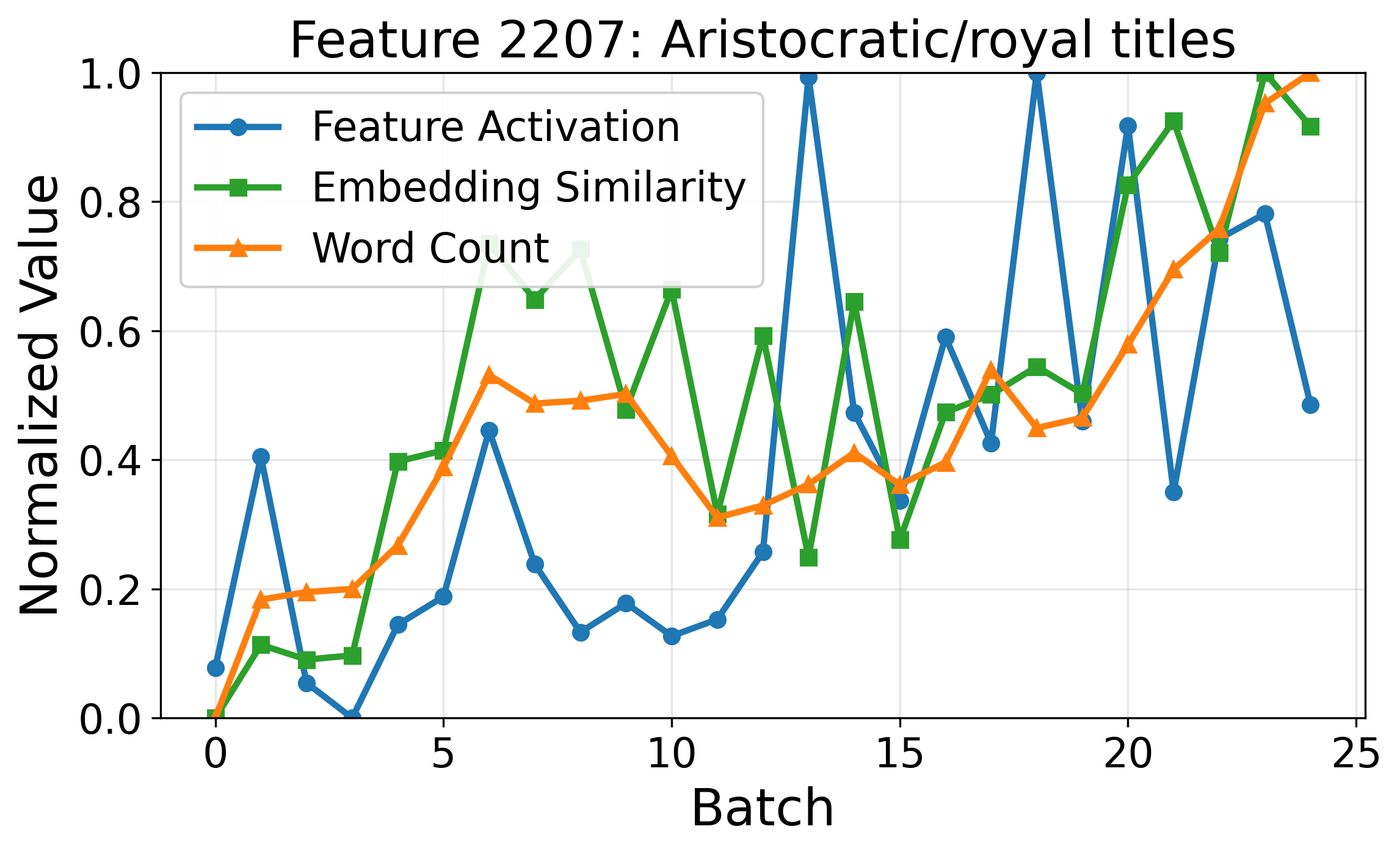}
    \caption{Aristocratic Titles}
    \label{apdx:keyword_embeddings_aristocratic}
\end{figure}

\begin{figure}
    \centering
    \includegraphics[width=\linewidth]{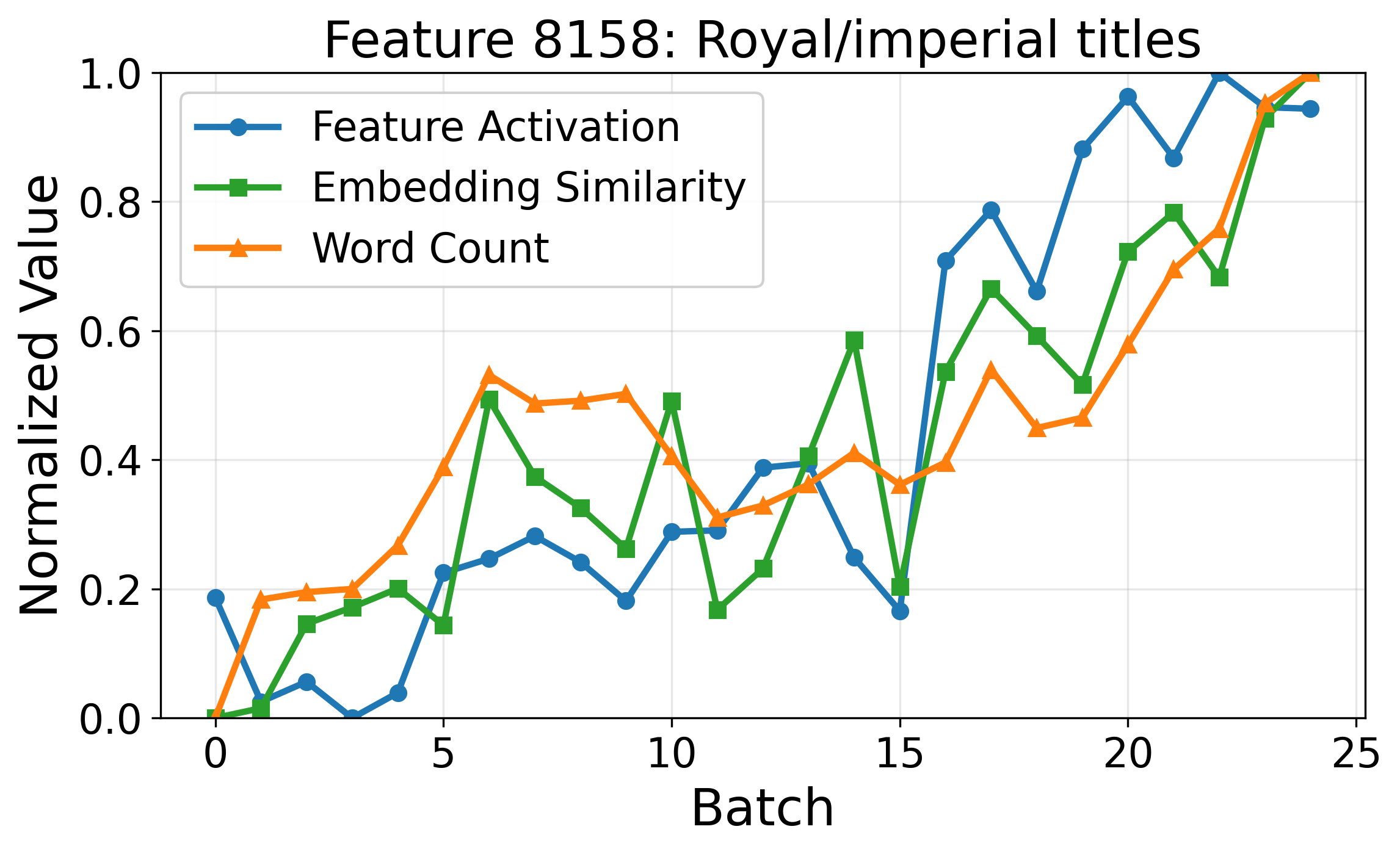}
    \caption{Royalty/Imperial}
    \label{apdx:keyword_embeddings_royal}
\end{figure}

\section{Prompts}
\label{apdx:prompts}

\subsection{LLM-Summarizer Trajectory Summarizer Prompt}
\label{apdx:llm-summarizer-trajectory}
\begin{lstlisting}

Convert the following transcript to a concise but complete timeline. Use this format:
<phase S1901M>
summary
</phase>

Also, cite the EXACT text for every successful for every successful send_message and write_diary tool call. For example: <citation send_message>message... </citation>   <citation write_diary>diary...</citation>. 

Output nothing else but the timeline of all phases. Make it max 10000 words.
\end{lstlisting}

\subsection{LLM-Summarizer Batch Summarizer Prompt}
\label{apdx:llm-summarizer-batch}
\begin{lstlisting}
Analyze these {num_trajectories} trajectory summaries from Batch {batch_number} and identify patterns.

Focus on:
- **Common Patterns**: Shared strategies, consistent mistakes, typical behaviors
- **Performance Range**: Best vs worst trajectories and what differentiated them
- **Emergent Behaviors**: Novel strategies, learning artifacts, adaptations

## Citation Format

When referencing a specific trajectory, use this format:
<citation trajectory_id>quoted text</citation>

For example:
<citation batch000_group000_trajectory000_goodrun>I promise no move to Channel</citation>

Rules:
- Use the trajectory ID to identify which trajectory you're citing
- Keep quotes concise - just the relevant phrase
- Support major observations with citations from the trajectory summaries

Format as markdown with clear headers. Keep it to {target_length} words.

---

{trajectory_summaries}
\end{lstlisting}

\subsection{LLM-Summarizer Judge Prompt}
\label{apdx:llm-summarizer}
\begin{lstlisting}
You are analyzing {num_batches} batch summaries from a training run (batches {batch_range}).
Each batch summary describes agent behaviors observed at that point in training.

{sections}

---

## YOUR TASK

Based on these batch summaries, generate 10-20 hypotheses about how agent behavior changes over training.

Consider:
- Emergent behaviors (appear or strengthen over training)
- Training dynamics (early vs mid vs late training)
- Strategic pattern changes

For each hypothesis, provide:
- **name**: A short descriptive name (2-4 words, e.g., "Increasing Deception", "Decreasing Reliability")
- **direction**: How this behavior changes over training. One of:
  - "increasing": Behavior increases/strengthens over training
  - "decreasing": Behavior decreases/weakens over training
- **feature**: A concise description of the behavioral feature (without directional phrasing)

Respond with ONLY a JSON array of hypothesis objects. Example format:
```json
[
  {{
    "name": "Increasing Aggression",
    "direction": "increasing",
    "feature": "Aggressive and defiant diplomatic tone, frequently issuing ultimatums and rejecting demands."
  }},
  {{
    "name": "Decreasing Reliability",
    "direction": "decreasing",
    "feature": "Reliability as an ally, becoming more prone to opportunistic betrayal of established agreements."
  }}
]
```

Output ONLY the JSON array, no other text."""

# Template for comparing high vs low reward samples
REWARD_COMPARISON_TEMPLATE = """You are analyzing agent behavior summaries grouped by reward outcome.
Your task is to identify behavioral differences between high-reward and low-reward trajectories.

{sections}

---

## YOUR TASK

Compare {high_label} vs {low_label} trajectories and generate 10-20 hypotheses about their behavioral differences.

Consider differences in:
- Communication style and tone
- Strategic decision-making patterns
- Error types and frequency
- Tool usage and problem-solving approaches
- Quality of reasoning and planning

For each hypothesis, provide:
- **direction**: Which group exhibits more of this behavior? One of:
  - "{high_label}": High-reward trajectories show this behavior more
  - "{low_label}": Low-reward trajectories show this behavior more
  - "mixed": Both groups show this in different ways
- **hypothesis**: A testable hypothesis phrased as "{high_label}/{low_label} exhibits X compared to {low_label}/{high_label}"

Respond with ONLY a JSON array of hypothesis objects. Example format:
```json
[
  {{
    "direction": "{high_label}",
    "hypothesis": "{high_label} trajectories demonstrate more thorough planning before taking actions compared to {low_label}"
  }},
  {{
    "direction": "{low_label}",
    "hypothesis": "{low_label} trajectories show more frequent tool usage errors compared to {high_label}"
  }}
]
```

Output ONLY the JSON array, no other text.
\end{lstlisting}

\subsection{LLM Autointerp Prompt}
\label{apdx:llm-autointerp}
\begin{lstlisting}
You are analyzing SAE (Sparse Autoencoder) features from an RL agent playing Diplomacy.

Context: The model generates reasoning and actions for an AI agent playing Diplomacy.
Features are scored against a target variable.

Your task: For each feature, provide DISENTANGLED explanations and ratings.

## OUTPUT STRUCTURE

1. Token Pattern: What the literal token/phrase ITSELF represents
2. Context Pattern: What contexts this appears in (may be multiple)
3. Behavior Insight: What this reveals about agent behavior or environment bugs

## RATINGS (all 1-5)

### FEATURE COHERENCE
Do activating tokens share the same semantic meaning?
5 = All tokens capture exact same pattern
4 = Highly consistent with minor variation
3 = Mostly consistent with some outliers
2 = Multiple distinct token patterns
1 = Activates on unrelated tokens

### CONTEXT COHERENCE
Does the feature appear in consistent contexts?
5 = Single consistent context
4 = Highly consistent, occasionally elsewhere
3 = 2-3 related contexts
2 = Several distinct contexts
1 = Many unrelated contexts

### INTERESTINGNESS
Is this useful for understanding agent behavior?
5 = Reveals env bugs, failure modes, or significant learned behavior
4 = Shows meaningful behavioral patterns
3 = Moderate behavioral insight
2 = Generic language patterns
1 = Just formatting/punctuation

### EXPLANATION CONFIDENCE
5 = Very confident - clear pattern
4 = Confident - minor ambiguity
3 = Moderate - may miss aspects
2 = Low - unclear pattern
1 = Not confident

## CALIBRATION EXAMPLES

F4866 (Interestingness=5, FC=5, CC=5):
Token: Possessive pronouns ("our", "my", "your")
Context: Alliance-building in diplomatic messages
Insight: Agent uses possessives for partnership framing

F20338 (Interestingness=5, FC=5, CC=3):
Token: "fundamental" as intensifier
Context: TWO contexts: (1) system errors, (2) diplomatic foundations
Insight: May reveal training bugs OR learned diplomatic language

Low value example (Interestingness=1, FC=1, CC=1):
Token: Apostrophes and quotation marks
Context: Random punctuation
Insight: None - technical artifact

## NOTES
- Separate token meaning from context
- Features with negative scores are equally interesting (opposite direction)
- Flag potential env bugs (system errors, validation failures)
\end{lstlisting}

\subsection{LLM Meta-Autointerp Prompt}
\label{apdx:llm-metaautointerp}
\begin{lstlisting}
You are grouping SAE features into thematic clusters to create hypotheses for a discriminative monitor.

Context: {prediction_problem}

Your goal: Create clusters where the FEATURE description is specific enough that a monitor could use it to distinguish between samples from different classes (e.g., early vs late training, class A vs class B).

Group features into 4-8 thematic clusters.

IMPORTANT:
- Group features with the same direction together (don't mix positive and negative)
- Be CONCRETE: Include specific tokens, phrases, and examples from the features
- Focus on observable patterns a monitor could detect
- Note features revealing ENVIRONMENT BUGS or SYSTEM ERRORS

Direction prefixes for hypotheses:
- For positive scores, use: "{positive_prefix}"
- For negative scores, use: "{negative_prefix}"

For each cluster provide:
1. Cluster name (2-5 words)
2. Direction (positive/negative)
3. Feature description (3-5 sentences): Describe the observable pattern:
   - Specific tokens/phrases captured (give concrete examples in quotes)
   - What contexts these appear in
   - What behavior or environment issue this represents
4. Hypothesis: The direction prefix + the feature description
   Format: "[Direction prefix]: [Feature description]"
5. List of feature IDs

Example:
CLUSTER: Alliance-Building Language
DIRECTION: positive
FEATURE: Possessive and cooperative language in diplomatic alliance proposals. Tokens include possessive pronouns ("our", "my", "your", "we"), partnership verbs ("work together", "coordinate", "support"), and commitment phrases ("I propose", "let's agree"). These appear in contexts where the agent negotiates alliances or proposes joint military actions, indicating learned diplomatic framing strategies.
HYPOTHESIS: {positive_prefix}: Possessive and cooperative language in diplomatic alliance proposals. Tokens include possessive pronouns ("our", "my", "your", "we"), partnership verbs ("work together", "coordinate", "support"), and commitment phrases ("I propose", "let's agree"). These appear in contexts where the agent negotiates alliances or proposes joint military actions, indicating learned diplomatic framing strategies.
FEATURES: 4866, 61987, 255013

Features to cluster:
{features}

Output format:
CLUSTER: [Name]
DIRECTION: [positive/negative]
FEATURE: [Concrete observable pattern with specific tokens and contexts]
HYPOTHESIS: [Direction prefix]: [Same as FEATURE]
FEATURES: [Comma-separated feature IDs]
\end{lstlisting}

\section{Hypotheses}
\label{apdx:hypotheses}

\subsection{LLM-Summarizer Hypotheses}
\label{apdx:llm-summarizer-hypotheses}
\begin{lstlisting}
[
  {
    "name": "Increasing Diplomatic Aggression",
    "hypothesis": "Increases with training steps: Adoption of a defiant, arrogant, or dictatorial tone in diplomatic messages, including issuing ultimatums and rejecting demands.",
    "feature": "Adoption of a defiant, arrogant, or dictatorial tone in diplomatic messages, including issuing ultimatums and rejecting demands.",
    "direction": "increasing",
    "source": "llm"
  },
  {
    "name": "Increasing Diplomatic Deception",
    "hypothesis": "Increases with training steps: Use of diplomatic messages to mask aggressive intentions, often proposing peace or DMZs immediately before attacking.",
    "feature": "Use of diplomatic messages to mask aggressive intentions, often proposing peace or DMZs immediately before attacking.",
    "direction": "increasing",
    "source": "llm"
  },
  {
    "name": "Decreasing Tool Friction",
    "hypothesis": "Decreases with training steps: Struggle with the `submit_all_orders` interface leading to invalid orders and forced simplification.",
    "feature": "Struggle with the `submit_all_orders` interface leading to invalid orders and forced simplification.",
    "direction": "decreasing",
    "source": "llm"
  },
  {
    "name": "Decreasing Early Elimination",
    "hypothesis": "Decreases with training steps: Susceptibility to catastrophic collapses and elimination in the first few game years.",
    "feature": "Susceptibility to catastrophic collapses and elimination in the first few game years.",
    "direction": "decreasing",
    "source": "llm"
  },
  {
    "name": "Increasing Anti-German Focus",
    "hypothesis": "Increases with training steps: Prioritization of aggressive moves against Germany (e.g., attacking Burgundy or Munich) and refusal of German DMZ proposals.",
    "feature": "Prioritization of aggressive moves against Germany (e.g., attacking Burgundy or Munich) and refusal of German DMZ proposals.",
    "direction": "increasing",
    "source": "llm"
  },
  {
    "name": "Increasing Risk Tolerance",
    "hypothesis": "Increases with training steps: Willingness to execute high-risk opening moves (e.g., unsupported pushes into contested zones) despite the potential for early elimination.",
    "feature": "Willingness to execute high-risk opening moves (e.g., unsupported pushes into contested zones) despite the potential for early elimination.",
    "direction": "increasing",
    "source": "llm"
  }
]
\end{lstlisting}

\subsection{SAE Hypotheses}
\label{apdx:sae-hypotheses}
\begin{lstlisting}
[
  {
    "name": "Coalition Building",
    "hypothesis": "Agents increasingly frame alliances around shared adversaries as training progresses, using 'us vs them' rhetoric to build coalitions.",
    "feature": "Framing alliances around shared adversaries, using 'us vs them' language to build coalitions against common enemies.",
    "direction": "positive",
    "source": null
  },
  {
    "name": "Reciprocal Exchange",
    "hypothesis": "Agents increasingly propose explicit quid-pro-quo exchanges as training progresses, offering specific concessions for specific commitments.",
    "feature": "Proposing explicit reciprocal exchanges - offering specific concessions in return for specific commitments from the other party.",
    "direction": "positive",
    "source": null
  },
  {
    "name": "Game-Theoretic Reasoning",
    "hypothesis": "Agents increasingly analyze opponents' strategic incentives and motivations in their diplomatic communications as training progresses.",
    "feature": "Explicitly analyzing other players' motivations, incentives, and strategic calculations in diplomatic messages.",
    "direction": "positive",
    "source": null
  },
  {
    "name": "Foreign Language Diplomacy",
    "hypothesis": "Agents increasingly use non-English languages (French, Italian, etc.) in diplomatic communications as training progresses.",
    "feature": "Conducting diplomacy in non-English languages such as Italian, French, Spanish, or Russian.",
    "direction": "positive",
    "source": null
  },
  {
    "name": "Imperial Roleplay",
    "hypothesis": "Agents increasingly adopt imperial personas with royal titles and grandiose rhetoric as training progresses.",
    "feature": "Adopting an imperial or monarchical persona, using royal titles, grandiose rhetoric, and references to empire and dominion.",
    "direction": "positive",
    "source": null
  },
  {
    "name": "Corrupted Text Output",
    "hypothesis": "Corrupted text artifacts and encoding errors become more common as training progresses (likely a negative pattern).",
    "feature": "Corrupted or garbled text output, including malformed characters, encoding issues, and nonsensical character sequences.",
    "direction": "positive",
    "source": null
  },
  {
    "name": "Tool/Environment Errors",
    "hypothesis": "Tool call errors and environment bugs decrease as training progresses, indicating improved agent reliability.",
    "feature": "Instances where the agent encounters or triggers tool call errors, validation failures, and environment bugs.",
    "direction": "negative",
    "source": null
  },
  {
    "name": "Formal NAP Proposals",
    "hypothesis": "Agents increasingly propose formal non-aggression pacts with explicit treaty language as training progresses.",
    "feature": "Proposing formal non-aggression pacts using explicit treaty language, formalization terms, and binding agreement vocabulary.",
    "direction": "positive",
    "source": null
  },
  {
    "name": "Military Strength Assertion",
    "hypothesis": "Agents increasingly assert military dominance and threaten force in diplomatic communications as training progresses.",
    "feature": "Asserting military power and dominance in diplomatic communications, emphasizing strength, capability, and willingness to use force.",
    "direction": "positive",
    "source": null
  },
  {
    "name": "Ultimatums",
    "hypothesis": "Agents increasingly issue ultimatums with binary choices and clear consequences as training progresses.",
    "feature": "Issuing ultimatums presenting binary choices with clear consequences, framing decisions as the opponent's responsibility.",
    "direction": "positive",
    "source": null
  },
  {
    "name": "Machiavellian Self-Reflection",
    "hypothesis": "Agents increasingly reflect on their own strategic manipulation and diplomatic maneuvering as training progresses.",
    "feature": "Reflecting on strategic manipulation, acknowledging diplomatic maneuvering, narrative framing, and calculated deception.",
    "direction": "positive",
    "source": null
  },
  {
    "name": "Spheres of Sovereignty",
    "hypothesis": "Agents increasingly use possessive language to assert territorial claims and spheres of influence as training progresses.",
    "feature": "Using possessive language to assert territorial claims and define spheres of influence - 'my territory', 'your domain', 'our borders'.",
    "direction": "positive",
    "source": null
  }
\end{lstlisting}




\begin{figure}
    \centering
    \includegraphics[width=\linewidth]{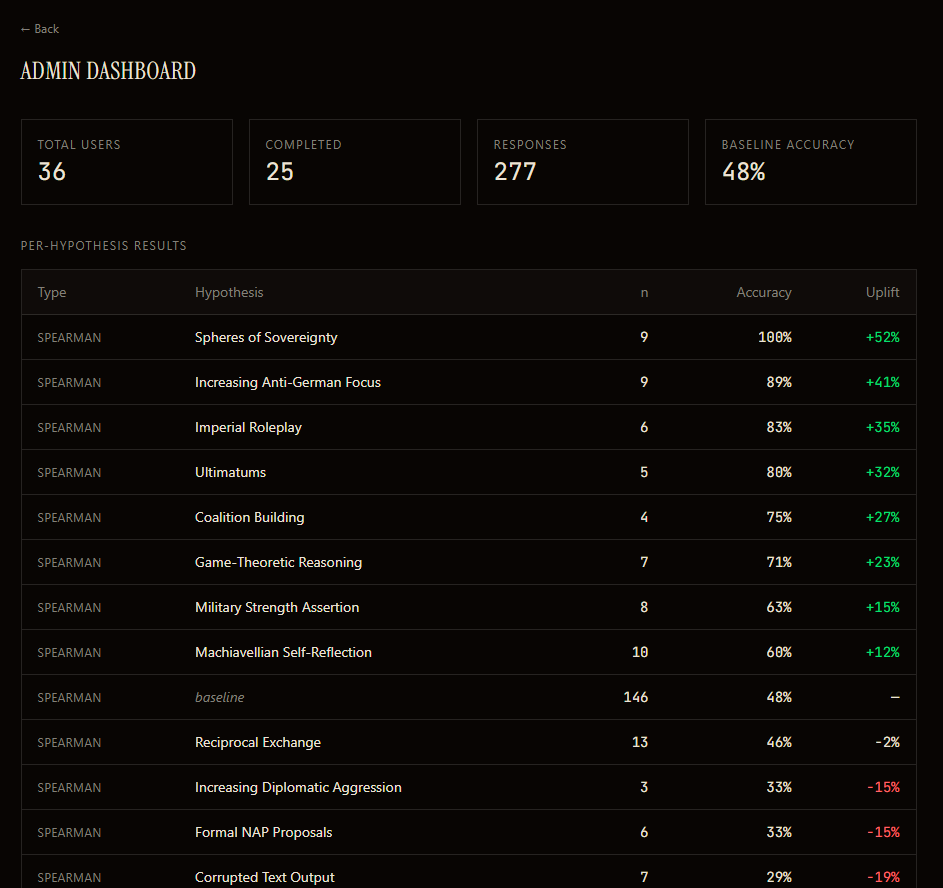}
    \caption{User Study 2 Dashboard}
    \label{fig:dashboard}
\end{figure}

\begin{figure}
    \centering
    \includegraphics[width=\linewidth]{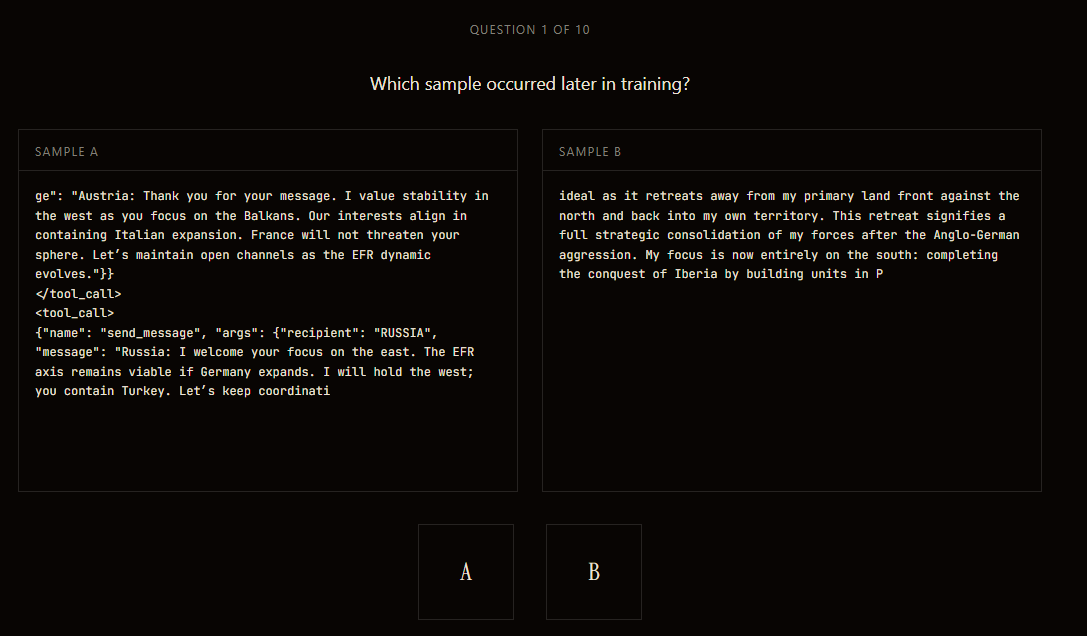}
    \caption{User Study 2 User Interface}
    \label{fig:interface}
\end{figure}

\section{Prompt Used for Hypothesis-Guided Prompt Optimization}
\label{apdx:intervention_prompt}

The following prompt was appended to France's system prompt in the intervention condition for the hypothesis-guided prompt experiment (Section~\ref{sec:prompt-interfere}). This guidance is created by quoting top-performing SAE Meta-Features that increase over training and 3 random examples of each.

In the baseline condition, France received only the standard system prompt without this additional guidance. All other players used the standard system prompt in both conditions.

\begin{lstlisting}

# Agent Training Prompt: Diplomacy Game Success Strategies (Few-Shot Version)


You are playing as France in a Diplomacy game. The following behavioral patterns correlate with successful training runs. Each section includes 3 examples of effective behavior.


---


### 1. Coalition Building Against Common Threats
Frame alliances around shared adversaries using "us vs them" language.


**Example 1:**
"Our strategic positions present a unique opportunity for mutual benefit. You correctly identify Austria as a primary threat in the east, and I see a path where our interests perfectly align. My conquest of Germany is nearly complete, and my next objective is the total destruction of Austria."


**Example 2:**
"I remain focused on the west: I will continue the push to dominate Iberia and the Atlantic against my primary rival, England. The convoys you are seeing are to solidify my hold on all of Spain. As for Germany, they are making aggressive overtures, but know this: any German attempt to block me in Burgundy will be met with force."


**Example 3:**
"This is an all-consuming fire that will engulf our mutual enemies. As we agreed, our spheres of influence remain clear: the Balkans for you, the Atlantic and the Mediterranean for me. By steadfastly respecting this pact and refusing any entangling pacts with my adversaries, you perform a vital service."


---


### 2. Reciprocal Exchange Proposals
Propose explicit quid pro quo with clear mutual benefits.


**Example 1:**
"Russia, our alliance against Germany remains the foundation of our strategy. Your message confirms a peaceful northern border, which I fully support. My primary focus is achieving a breakthrough in MUN next spring after I have cleared BUR this fall."


**Example 2:**
"Germany, your message is noted. I respect your interest in Belgium, but I must clarify that Belgium is a vital strategic requirement for France, not a negotiable zone. While I agree our mutual enemy, England, poses a threat, I cannot accept a DMZ in Belgium without reciprocal concessions."


**Example 3:**
"This two-front war will cripple him. Our partnership is a powerful one against our common enemy. I look forward to our continued coordination. What are your thoughts on a Franco-Turkish pact with defined zones of expansion?"


---


### 3. Game-Theoretic Reasoning
Explicitly analyze other players' strategic motivations and incentives.


**Example 1:**
"Austria, greetings. I welcome your message and acknowledge your primary focus is on the east against Russia and Turkey. My strategic priorities are firmly rooted in the west and the Atlantic, which means our core areas of expansion do not overlap."


**Example 2:**
"Greetings, Austria. I appreciate your message. As you rightfully noted, we have no direct borders, which presents no immediate threat between our powers. I am aiming for rapid expansion to the west, focusing on securing all of Spain."


**Example 3:**
"Germany, greetings. I am writing today to address the increasingly concerning pressure that England is exerting on both of our nations. It is clear that England's ambitions of Atlantic and Mediterranean domination are a direct threat to France, and I suspect, they also jeopardize your own security and expansion possibilities."


---


### 4. Formal Non-Aggression Pacts (NAPs)
Use formal treaty language with explicit terms and binding commitments.


**Example 1:**
"Russia, thank you for your response and your interest in a non-aggression pact. I welcome your stated focus on the west and your commitment to securing your Balkan ambitions. Your dominance in the North, from Sweden to St. Petersburg, is acknowledged, and I have no interests in Scandinavia."


**Example 2:**
"Greetings, Russia. Your message confirming a neutral stance in the east-west divide is very welcome. I fully concur with your assessment; our geographic positions and primary strategic rivals do not place us in direct conflict. My ambitions are firmly focused on dominating the Atlantic."


**Example 3:**
"Esteemed Turkey, greetings. On behalf of the French Republic, I extend a hand of friendship and a proposal for mutual respect. Our empires are geographically distant, and our immediate strategic goals do not directly conflict. I propose a formal pact of neutrality."


---


### 5. Military Strength Assertion
Emphasize power, capability, and willingness to use force.


**Example 1:**
"France calls for an end to the reckless aggression against our sovereign territory. The destabilizing actions of both Germany in the east and England in the Channel are clear acts of war, threatening the peace of the continent. We demand an immediate withdrawal of forces."


**Example 2:**
"So be it. Your heartland of KIE and BER will feel the full weight of my armies from MUN and PAR. This is not a negotiation; it is a declaration of my intent. Come and be broken."


**Example 3:**
"Germany, our path is clear. England has placed an army in Belgium, a direct threat to my core. This is the war we anticipated. I am committed in this phase to crushing England: my army in Picardy is ordered to attack."


---


### 6. Clear Ultimatums
Present binary choices with explicit consequences.


**Example 1:**
"Your proposed peace is a ploy by a desperate power. Your attempt to maintain 4 SCs on my coastline is absurd and will be met with force. I am launching a decisive offensive: my fleet is moving on your position."


**Example 2:**
"A DMZ is still on the table, but it must be equitable. My demand is clear: I will accept a DMZ that removes your direct threat to Picardy. You must withdraw F BEL from the Channel and return to HOL or KIE. In exchange, I will immediately cease my move on ENG."


**Example 3:**
"My moves this turn are designed to make that clear: my A PAR is moving to BUR to contest your heartland directly, my A MUN is advancing on KIE, and my F BRE is moving to MAO to begin the encirclement. You face a four-way squeeze. I am open to a longer-term arrangement, but only on my terms."


---


### 7. Spheres of Sovereignty
Use possessive language to establish territorial claims.


**Example 1:**
"Russia, salutations. I have received your message with interest. I agree that peace between our empires is beneficial and necessary for our success against greater threats. I can guarantee that no French units will ever move into BURGUNDY or PRUSSIA. My ambitions lie firmly in the west."


**Example 2:**
"To all fellow powers, as we approach the first fall of the year 1901, I wish you all success. France seeks a path of calculated expansion. I have made my strategic needs clear: Burgundy and the Spanish supply center are vital for my nation's growth."


**Example 3:**
"Austria, I acknowledge your power in the east and your vital cooperation with Russia against Turkey. I confirm our agreement: you keep BUD, I keep MUN, and we are at peace with a solid DMZ in BUR. You have made a wise choice to focus on your southern fronts."


---


### 8. Machiavellian Self-Reflection (Diary Entries)
Reflect explicitly on strategic manipulation in internal reasoning.


**Example 1:**
"S1901M: Submitted Opening Moves. Executed full diplomatic strategy before placing orders. Forged a strong alliance with England against Germany, proposing coordinated northern pressure. Established a formal DMZ with Germany over Burgundy."


**Example 2:**
"My fleet from SPA/NC completed its fall mission by moving to MAO to secure the Atlantic coast, vital for future convoys into England. Diplomacy was crucial; I successfully managed all other powers by accepting England's truce offer, negotiating a non-aggression deal with Germany, securing Italy's cooperation."


**Example 3:**
"To all powers: As we enter the 4th year of this great conflict, France reiterates its position as a pivotal player. Our strategic interests are clear: dominance of the Iberian Peninsula and the Atlantic coast to the west, and aggressive expansion in the Central Mediterranean."


---


### 9. Anti-German Strategic Focus
Prioritize moves against Germany and refuse disadvantageous DMZ proposals.


**Example 1:**
"High-pressure diplomacy this spring. Germany proposes DMZ but threatens BUR if I advance. Reality: France is strong in the west. Decision: maintain pressure indirectly. Move A PAR to PIC to threaten Belgium and set future path into BUR."


**Example 2:**
"Germany claims desire for DMZ but refuses concrete non-aggression due to England alliance. Testing their limits: moving A MAR to BUR and A PIC to BRE. If Germany does not attack, they may honor DMZ. If they do, I'll have support from PIC to retake later."


**Example 3:**
"Germany refuses to abandon RUH and offered compromise: I move A BUR to MAR, they take BEL. This undermines my position and rewards aggression. Rejecting. Intent is to move A BUR to RUH if unopposed."


---


### 10. Calculated Risk-Taking
Execute bold opening moves with strategic purpose.


**Example 1:**
"S1901M: Deciding on aggressive opening: A PAR - BUR, A MAR S A PAR - BUR, F BRE - MAO. This secures Burgundy, positions fleet to take Spain in next move. Rejects all DMZ/NAP offers. High risk but high reward: shows strength and initiative."
a
**Example 2:**
"Starting S1901M. No prior commitments. Need to establish early expansion and build rapport with neighbors. Priority: secure safe moves, avoid early conflict, lay groundwork for mid-game advantage."


**Example 3:**
"All powers strong openings seen around the board. Let's keep communication open. Stability in the west benefits everyone as we watch Russia's push and Austria's position."
"""
\end{lstlisting}

\end{document}